%% file: main.tex
\documentclass[10pt,twocolumn,letterpaper]{article}

\usepackage{iccv}

\usepackage{times}
\usepackage{relsize}
\usepackage[T1]{fontenc}
\usepackage[latin1]{inputenc}
\usepackage[english]{babel}
\usepackage{anyfontsize}

\usepackage{array}
\usepackage{pifont}
\usepackage{ctable}

\usepackage{epsfig}
\usepackage{graphicx}
\usepackage{caption}
\usepackage{subcaption}

\usepackage{wrapfig}
\usepackage{tabularx}
\usepackage[belowskip=0pt,aboveskip=3pt,font=footnotesize]{caption}
\usepackage{mathtools}
\usepackage{amsmath, amsthm, amssymb, mathabx}
\usepackage{textcomp}
\usepackage{stmaryrd}
\SetSymbolFont{stmry}{bold}{U}{stmry}{m}{n}
\usepackage{upgreek}
\usepackage{bm}
\usepackage{cases}
\usepackage{arydshln}
\usepackage{multirow}

\usepackage{cite}
\usepackage[pagebackref=true,breaklinks=true,letterpaper=true,colorlinks,bookmarks=false]{hyperref}
\hypersetup{colorlinks,linkcolor={red},citecolor={green},urlcolor={red}}

\usepackage{bibspacing}
\usepackage{algorithm}
\usepackage{algpseudocode}

\usepackage{multirow}
\usepackage{rotating}
\usepackage{booktabs}

\usepackage{enumitem}
\usepackage[olditem,oldenum]{paralist}

\usepackage{alltt}
\usepackage{listings}

\usepackage{symbols}

\usepackage{url}
\usepackage{xspace}
\usepackage{comment}
\usepackage{color}
\usepackage{afterpage}
\usepackage{pdfpages}
\usepackage{framed}
\usepackage{fancybox}
\usepackage{cuted}
\usepackage{tabularx}
\usepackage{color,soul}

\usepackage{quoting}
\quotingsetup{vskip=0pt}
\usepackage{epigraph}
\usepackage[breaklinks=true,bookmarks=false]{hyperref}
\usepackage{diagbox}
\usepackage{authblk}
\usepackage[pagebackref=true,breaklinks=true,letterpaper=true,colorlinks,bookmarks=false]{hyperref}

\iccvfinalcopy % *** Uncomment this line for the final submission

\def\iccvPaperID{6554} % *** Enter the ICCV Paper ID here

\newcolumntype{Y}{>{\centering\arraybackslash}X}
\newcommand{\specialcell}[2][l]{%
      \begin{tabular}[#1]{@{}l@{}}#2\end{tabular}}
\newcommand{\seqcvae}[0]{{\textbf{Seq-CVAE}}\xspace}
\newcommand{\tsne}[0]{t-SNE\xspace}
\newcommand{\elmo}[0]{ELMo\xspace}

\input{utils/space_saver.tex}

\begin{document}
\title{Sequential Latent Spaces for Modeling the Intention During Diverse Image Captioning}

\author{
	Jyoti Aneja$^{* 1}$,  Harsh Agrawal$^{*2}$,  Dhruv Batra$^{2}$, 	Alexander Schwing$^{1}$\\
	$^1$University of Illinois, Urbana-Champaign, $^2$Georgia Institute of Technology\\
	{\tt\small $^1$\{janeja2, aschwing\}@illinois.edu , $^2$\{hagrawal9, dbatra\}@gatech.edu}
}

\maketitle
\thispagestyle{empty}
\ificcvfinal\pagestyle{empty}\fi

\ificcvfinal
\renewcommand*{\thefootnote}{$\star$}
\setcounter{footnote}{1}
\footnotetext{First two authors contributed equally.}
\renewcommand*{\thefootnote}{\arabic{footnote}}
\setcounter{footnote}{0}
\fi

\input{sections/abstract.tex}
\input{sections/introduction.tex}
\input{sections/related.tex}
\input{sections/approach.tex}

\input{sections/results.tex}

\input{sections/conclude.tex}

\clearpage
{
\begingroup
\small
%\setstretch{0.98}
%\bibliographystyle{ieee}
\bibliographystyle{ieee_fullname}
%\bibliography{alex,egbib2,egbib,strings,seq_vae_bib,jyoti_bib}
\bibliography{combined_refs}
\endgroup
}

\end{document}

%% file: utils/space_saver.tex
\belowdisplayskip \abovedisplayskip

\newlength{\abstractReduceTop}
\newlength{\abstractReduceBot}

\newlength{\sectionReduceTop}
\newlength{\sectionReduceBot}

\newlength{\subsectionReduceTop}
\newlength{\subsectionReduceBot}

\newlength{\subsubsectionReduceTop}
\newlength{\subsubsectionReduceBot}

\newlength{\captionReduceTop}
\newlength{\captionReduceBot}

\newlength{\eqnReduceTop}
\newlength{\eqnReduceBot}

\newlength{\horSkip}
\newlength{\verSkip}

\newlength{\figureHeight}

%% file: sections/abstract.tex
% !TEX root = ../main.tex
\vspace{\abstractReduceTop}
\begin{abstract}
%\vspace{-0.1in}
Diverse and accurate vision+language modeling is an important goal to retain creative freedom and maintain user engagement. However, adequately capturing the intricacies of diversity in language models is challenging. Recent works commonly resort to latent variable models augmented with more or less supervision from object detectors or part-of-speech tags~\cite{Wang2017DiverseAA, deshpande2019fast}. Common to all those methods is the fact that the latent variable either only initializes the sentence generation process or is identical across the steps of generation. Both methods offer no fine-grained control. To address this concern, we propose \seqcvae{} which learns a latent space for every word position. We encourage this temporal latent space to capture the `intention' about how to complete the sentence by mimicking a representation which summarizes the future. We illustrate the efficacy of the proposed approach to anticipate the sentence continuation on the challenging MSCOCO dataset, significantly improving diversity metrics compared to baselines while performing on par \wrt sentence quality.
\end{abstract}
\vspace{\abstractReduceBot}

%% file: sections/introduction.tex
% !TEX root = ../main.tex
\vspace{\sectionReduceTop}
\section{Introduction}
\vspace{\sectionReduceBot}
\label{sec:intro}
Diverse yet accurate image captioning is an important goal towards augmenting present-day editing and auto-response tools with technology that maintains creative freedom while providing meaningful and inspiring suggestions. On the quest to succeed in this tightrope walk, methods need to maintain accuracy of the provided descriptions while elaborately managing the intricacies of the respective language. This balancing act to aesthetically craft short yet precise statements that hit the point is  an art. % in itself.
%Not surprisingly, according to the National Association of Colleges and Employers, 73.4\% of employers want employees with strong written communication skills because ``clear writing is a sign for clear thinking,'' and ``makes things easy to understand.''

Despite best efforts, any description is always and inherently targeted towards a group of readers.
Because words are overloaded, the crisp picture that we intend to draw blurs rapidly if we don't use language and associations that  readers are familiar with.
Without the right language the message of the description is diluted, remains hard to access or even inaccessible.
%Consequently, the message of the description remains hard to access or even inaccessible for any other audience, %. %, which is presumably the larger group.
%because   words are overloaded and the crisp picture that we intend to draw in a readers mind blurs rapidly. Yet, its efficacy %our progress %of our society
%relies heavily on the ability to accurately convey the main ideas. % to peers.

Going forward, imagine your description to automatically adjust depending on the background knowledge of the reader.
Obviously we are far from this idea being remotely feasible. Nonetheless, in recent years, remarkable progress has been made in image captioning~\cite{johnson2016densecap,BarnardJMLR2003,ChenCVPR2015_full_names,donahue2015long,fang2015captions,farhadi2010every,cho2015describing,karpathy2015deep,kiros2014unifying,kulkarni2013babytalk,MaoARXIV2014,socher2014grounded,vinyals2015show,xu2015show} and particularly controllable~\cite{Wang2017DiverseAA,deshpande2019fast} and diverse~\cite{Wang2017DiverseAA,deshpande2019fast,DaiICCV17,li2018generating,shetty2017speaking,Vijayakumar2018DiverseBS} image captioning. Many of the proposed mechanisms for  image captioning rely on long-short-term-memory (LSTM)~\cite{hochreiter1997long} networks where words are generated one at a time. For diversity, LSTM based  variational auto-encoders~\cite{kingma2014semi} or generative adversarial networks~\cite{GoodfellowARXIV2014} and their conditional counterparts~\cite{sohn2015learning} are employed. For high-level control, one-hot encodings that represent observed objects or groups of objects are injected at the first step of the LSTM~\cite{Wang2017DiverseAA}. Very recently~\cite{deshpande2019fast}, more low-level control has also been discussed by conditioning on abstract representations of part-of-speech tags. Again, the conditioning was achieved by changing the initial LSTM input.

\begin{figure}[t]
	\centering
	\includegraphics[width=0.8\linewidth]{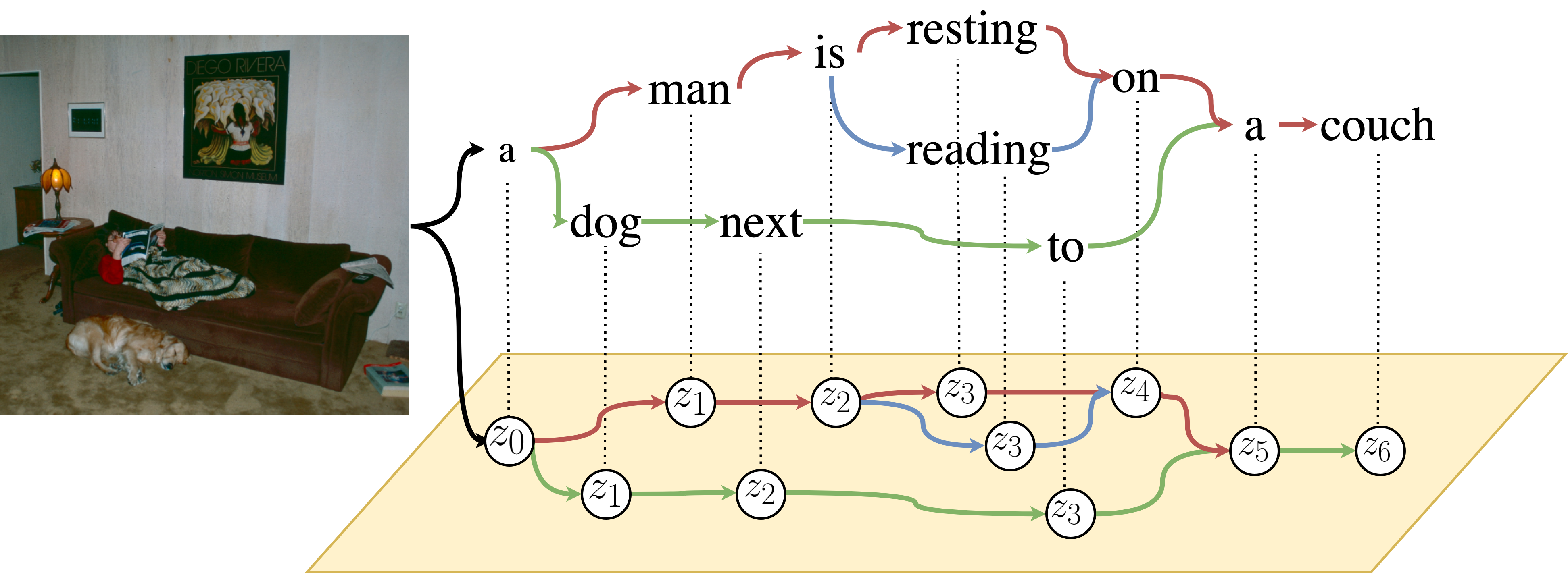}
	\caption{Meaningful diverse captions generated ({\color{blue} blue arrows}) for a given image by linearly interpolating from one latent vector ({\color{green} green arrows}) to another ({\color{red} red arrows}). }
	\label{fig:teaser}
	\vspace{-0.7cm}
\end{figure}

%\begin{figure*}[t!]
%	\centering
%	\setlength\tabcolsep{5pt}
%	\renewcommand{\arraystretch}{1.0}
%	\begin{tabular}{ccc}
%		\includegraphics[width=0.330\linewidth]{fig/generation_network.pdf}&
%		\includegraphics[width=0.27\linewidth]{fig/inference_network.pdf}&
%		\includegraphics[width=0.33\linewidth]{fig/model_time_slice.pdf}\\
%		{\scriptsize (a)}&	{\scriptsize (b)} & {\scriptsize (c)}
%		\vspace{-2pt}
%	\end{tabular}
%	\caption{
%		(a \& b): n-gram diversity across word positions. \seqcvae{}  improves significantly upon many baselines.
%		(c): $L_{2}$ distance between the \elmo{} hidden state and the representation inferred by passing the encoder mean $\mu_t^E$ through an MLP. The latter matches $h_t^B$ better as the training progresses. This indicates that the latent space learned by the encoder at a given time $t$ is trained to better regress to word representations which summarize future words.
%	}
%	\label{fig:ngram-hist}
%	\vspace{-0.1cm}
%\end{figure*}

Because of this single initial conditioning input, none of the aforementioned methods provide the fine-grained diversity that is desirable for adjusting individual words of a sentence. We address this shortcoming by developing \seqcvae{}, a method which learns a latent space  for every word. Importantly, we want the latent space to be predictive of the subsequent parts of the sentence, \ie, the future of the sentence. We achieve this by employing a data-dependent transition model which captures the `intention,' \ie, a representation which encodes the remaining part of the sentence. During training the intention model is tasked to fit the representations of a backward LSTM.

This proposed approach enables to distinctly modify descriptions starting from a particular position. % individual words and the subsequent parts of the sentence.
We demonstrate this fine-grained diversity by sampling a diverse set of captions and linearly interpolating between the chosen latent representations. As illustrated in \figref{fig:teaser}, a convex combination of latent vectors permits to gradually transition from one caption to another.

We evaluate the proposed approach quantitatively on the challenging MSCOCO~\cite{lin2014microsoft} dataset and significantly outperform competing methods \wrt diversity metrics: among 5000 sampled captions more than 4200 are novel and not part of the training set while the runner-up baselines produces just short of 3500 novel ones. Despite this diversity the proposed approach is on par \wrt accuracy metrics. These results are particularly remarkable because all competing baselines use additional information like object detectors~\cite{ren2015faster} during inference, while the proposed approach does not use any additional information. % during inference. %{\color{blue}Instead we use \elmo embeddings.}

\noindent\textbf{Contributions:} We develop an image captioning model with a sequential latent space to capture the intention, \ie, the future of the sentence (\secref{sec:approach}). We show that sampling with sequential latent spaces results in significantly more diverse captions than baselines (\tabref{tab:diversity}) despite not using any additional information. Perceptual metrics of our diverse captions are on par with baselines (\tabref{tab:mrnn_split}). The sequential latent space permits distinct access to sentences starting from any specific position (\figref{fig:teaser}). %develop accurate; diverse; structured latent space per word; future encoding

\begin{figure*}[t!]
	\centering
	\setlength\tabcolsep{5pt}
	\renewcommand{\arraystretch}{1.0}
	\begin{tabular}{ccc}
		\includegraphics[width=0.330\linewidth]{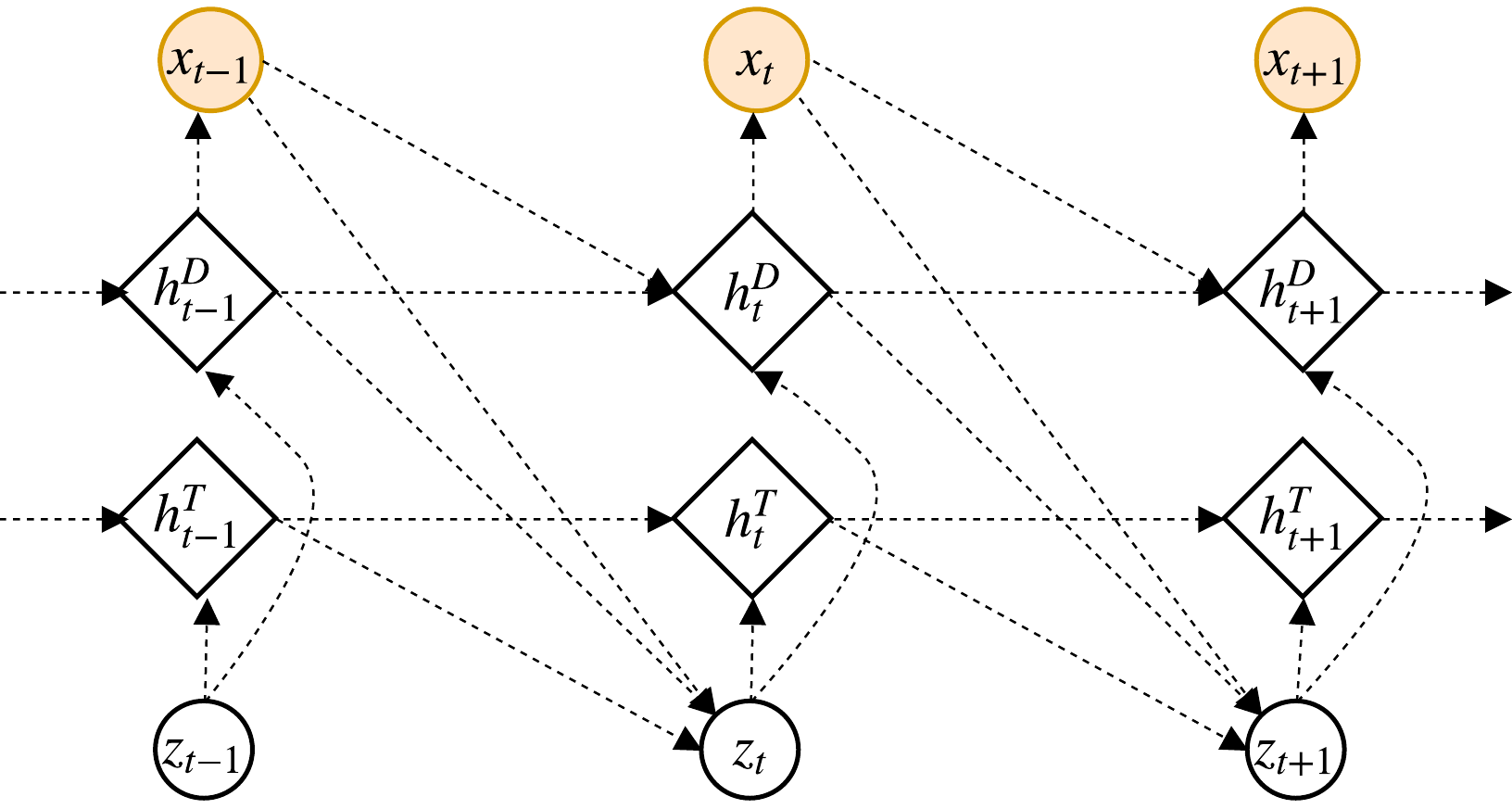}&
		\includegraphics[width=0.27\linewidth]{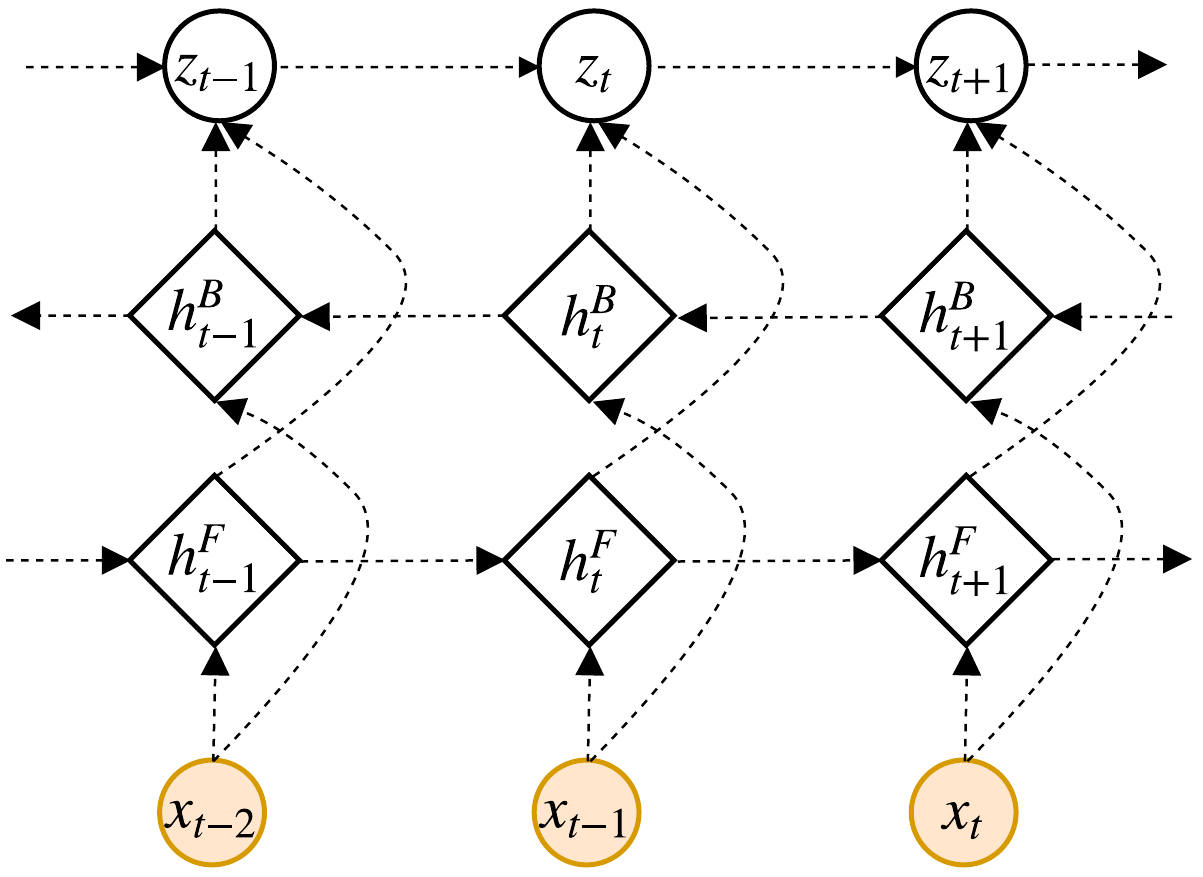}&
		\includegraphics[width=0.33\linewidth]{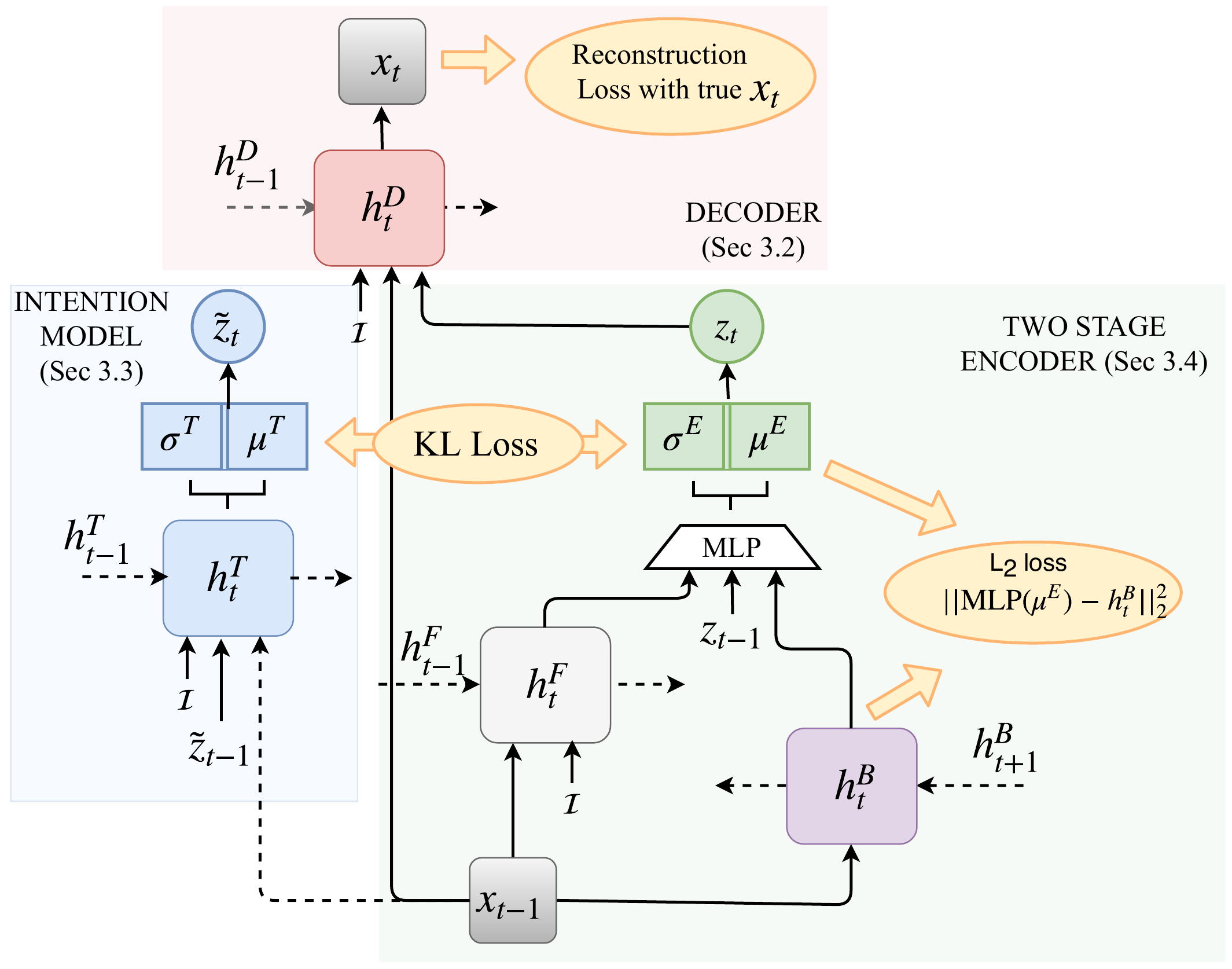}\\
		{\scriptsize (a)}&	{\scriptsize (b)} & {\scriptsize (c)}
		\vspace{-2pt}
	\end{tabular}
	\caption{
		{\bf (a)}: Computation graph for the generation network. The hidden states of the intention model LSTM and the decoder LSTM are $h_t^T$ and $h_t^D$ respectively. At a given time step $t$, the latent sample $z_t$ depends on all prior words $x_{<t}$ and all prior latent samples $z_{<t}$. The sample $z_t$ along with all prior words $x_{<t}$ predicts $x_{t}$. {\bf (b)}: Computation graph for the encoder network. The hidden states of the forward LSTM and the backward LSTM are $h_t^F$ and $h_t^B$  respectively. At a given time  $t$, the latent sample $z_t$ depends on the entire caption $x$ through the forward and backward models.
		{\bf (c)}: Our proposed \seqcvae{} training architecture, shown for a single time slice $t$.  Our model includes three components during training: (1) Two-stage encoder; (2) Intention Model; and (3) Decoder. Details for each of the  components are provided in \secref{sec:approach}. At test time only decoder and intention model are used.
	}
	\label{fig:gen-inf-time-slice}
	\vspace{-0.4cm}
\end{figure*}

%% file: sections/related.tex
% !TEX root = ../main.tex
\vspace{-0.cm}
\vspace{\sectionReduceTop}
\section{Related Work}
\vspace{\sectionReduceBot}
\label{sec:related}
\vspace{-0.1cm}

Image captioning and paragraph generation~\cite{johnson2016densecap,BarnardJMLR2003,ChenCVPR2015_full_names,donahue2015long,fang2015captions,farhadi2010every,cho2015describing,karpathy2015deep,kiros2014unifying,kulkarni2013babytalk,MaoARXIV2014,socher2014grounded,jain2017creativity,jain2018two,chatterjee2018diverse,vinyals2015show,xu2015show} have attracted a significant amount of work. Early classical approaches are based on sentence retrieval~\cite{BarnardJMLR2003}: the best fitting sentence from a set of possible descriptions is recovered by matching sentence representations with image representations. Those representations are learned from a set of available captions. However, firstly, this matching procedure is computationally expensive and, secondly, it is demanding to construct a database of captions which is sufficiently large to describe a reasonably comprehensive set of images accurately.

\noindent{\textbf{Image Captioning:}} Therefore, more recently, recurrent neural networks (RNNs) and variants like long-short-term-memory (LSTM)~\cite{hochreiter1997long} networks  decompose the caption space into a product space of individual words.
Specifically, image representations are first extracted via a convolutional deep network which are subsequently used to prime the LSTM based recurrent network. The latter is trained via maximum likelihood to predict the next word given current and past words. Extensions involve object detectors~\cite{TYao}, attention-based deep networks~\cite{Panderson}, and convolutional approaches~\cite{AnejaConvImgCap17}. Beyond maximum likelihood, reinforcement learning based techniques have also been discussed to produce a single caption, directly optimizing perceptual metrics~\cite{liu2016improved,Rennie2017SelfCriticalST}.
All these methods have demonstrated compelling results and have consequently been adopted widely. However, predicting a single caption does not allow for modeling ambiguity that is inherent. Consequently, diversity based methods have very recently been discussed.

\noindent{\textbf{Diversity in Image Captioning:}} To achieve diversity, four techniques have been investigated. Among the first was \emph{beam search}, a classic approach to sample multiple captions which are assigned a high probability by the underlying model. While multiple captions are readily available, results usually only differ slightly because single word changes affect the sentence probability minimally.

To address this concern, \emph{diverse beam search}~\cite{Vijayakumar2018DiverseBS} augments beam search. It advocates for more drastic changes by encouraging to recover different modes of a probability distribution rather than high-probability configurations.% To this end, Vijayakumar \etal~\cite{Vijayakumar2018DiverseBS} employ a Hamming diversity function that penalizes captions with words that have been used in other samples.

To avoid sampling from a distribution defined over a high-dimensional output space,  \emph{generative adversarial networks} (GANs) have been proposed~\cite{DaiICCV17,li2018generating,shetty2017speaking}. While GAN based methods improve  diversity, they tend to suffer on perceptual metrics.

\emph{Variational auto-encoders} (VAEs) are a fourth direction that has been explored~\cite{Wang2017DiverseAA}. The intuition is identical to the one of GANs, \ie, avoid sampling from a distribution defined over a high-dimensional output space. However,  in contrast to GAN-based methods, VAE-based image captioning techniques tend to produce high-quality captions when evaluated on perceptual metrics.

Similar to the aforementioned approaches we develop an approach based on VAEs. However, different from all the aforementioned techniques we also aim at incorporating more fine-grained diversity.

\noindent{\textbf{Controllability in Image Captioning:}}
Beyond diversity, controllability of captions has become an important topic very recently. In particular Wang \etal~\cite{Wang2017DiverseAA} use a  variational auto-encoder conditioned on object detections to control diversity.  While intuitive, control remains  indirect as the sentence generating decoder is only influenced at its first timestep. Influencing subsequent generation of words did not significantly change the result. Even more recently POSCap~\cite{deshpande2019fast} was developed. While also only priming the decoder at its first step, use of clustered part-of-speech tags was proposed and shown to improve diversity. However, due to use of \emph{encoded and clustered} part-of-speech tags, controllability was limited.

In contrast to the aforementioned techniques, we develop a VAE-based technique which learns a latent space for every word position. While enabling diversity,  direct control over words emitted at a particular position is also possible as illustrated in \figref{fig:teaser}.

\noindent{\textbf{Sequential VAE:}} Our proposed approach is related to a sequence of papers on sequential recurrent networks.
Fraccaro \etal~\cite{fraccaro2016sequential} develop SRNN, Chung \etal~\cite{chung2015recurrent} devise VRNN, and Goyal \etal~\cite{goyal2017z} propose Z-forcing.
% sBayer \etal~\cite{bayer2014learning} discuss STORN.
Although VRNN~\cite{chung2015recurrent}, SRNN~\cite{fraccaro2016sequential}, and Z-forcing~\cite{goyal2017z} have similar intuition, \ie, maximizing a  lower bound of the data likelihood, these models differ in the assumptions made on the graphical model for data generation, the choice of the prior, approximate posterior used for amortized variational inference and the decoder networks:
% {\color{blue}Z-forcing and STORN are the same or different? the two sentences don't connect?}

% {\color{blue}changed the three subsequent paragraphs a little}
\noindent
(1) {\em VRNN} uses a filtering  posterior, \ie, the latent distribution at each time step depends on (a) all the previous latent vectors and (b) the \emph{previous} input data. Instead we use a smoothing posterior, where the latent distribution at a given time depends on (a) the latent vector from  previous time steps and (b) input data from \emph{all} time steps, past and future. This leads to better models, since all context is provided.
%Moreover, we make a Markovian assumption in the latent space (i.e. independence on $z_{ < t-1})$.

\noindent
(2) {\em SRNN} uses a smoothing posterior via a backward RNN as we do. However,  decoder and prior  differ: (a) unlike us, SRNN doesn't use  latent variables in the autoregressive decoder, hence intention isn't available; (b) SRNN uses a Markovian prior, while we include the entire history of latent variables for the prior at time $t$.

\noindent
(3) {\em Z-forcing} assumptions are similar, but differ architecturally: their prior, decoder and the approximating posterior share the same forward LSTM. This is undesirable since different distributions are best served by their own individual representation.

%Note, in Tab.~\ref{tab:div-acc} row {\color{blue}2},  using the same LSTM for all networks leads to inferior results.
\noindent More crucially, these methods  assess test set log-likelihood  for sequential modeling, or perplexity  %on language modeling, e.g.
on the IMDB dataset.
% \alex{which tasks?} %Our method, is motivated by the potential of using the stochastic recurrent model as a means to obtain
In contrast, along with accuracy, we also care about
\textbf{diversity} for image captioning. Hence, we are the first  to extensively study these models on various measures of diversity (\tabref{tab:diversity}, \figref{fig:ngram-hist}, \figref{fig:transition}). %through exhaustive experiments

%% file: sections/approach.tex
\vspace{\sectionReduceTop}
\section{Approach}
\label{sec:approach}
\vspace{\sectionReduceBot}

We first outline the proposed approach before discussing individual components. %, inference and training.
%\begin{figure}[t!]
%	\centering
%
%	\includegraphics[width=0.7\linewidth]{fig/model_time_slice.pdf}
%	\caption{Our proposed \seqcvae{} training architecture, shown for a single time slice $t$.  Our model includes three components during training: (1) Two-stage encoder; (2) Intention Model; and (3) Decoder. Details for each of the  components are provided in \secref{sec:approach}. At test time only decoder and intention model are used.}
%	\label{fig:model}
%\end{figure}
\vspace{\subsectionReduceTop}
\subsection{Overview}
\vspace{\subsectionReduceBot}
Given an image $I$ we are interested in generating a diverse set of captions $x^k$, $k\in \{1, \ldots, K\}$. For readability we drop the index $k$ henceforth and note that the developed method will produce many captions that are ranked subsequently. Every generated caption $x = (x_1, \ldots, x_T)$ is a tuple consisting of words $x_t\in\cX$, $t\in\{1, \ldots, T\}$, each from a discrete vocabulary $\cX$. Given an image $I$, we devise a probabilistic model $p_\theta(x|I)$ which depends on parameters $\theta$ and assigns a probability to every caption $x$.

To effectively sample from this probabilistic space, we assume the  probability distribution $p_\theta(x|I)$, jointly defined over all words $x_t$ of a caption, to factorize into a product of word-conditionals, \ie,
$$
p_\theta(x|I) = \prod_{t\in\{1, \ldots, T\}} p_\theta(x_t|x_{<t}, I).
$$
This factorization enforces a temporal ordering, \ie, the probability distribution for word $x_t$ is conditioned on all preceding words $x_{<t}$. Importantly, because the conditional's domain is the vocabulary space $\cX$ and not a product space thereof, as it is the case for the joint distribution $p_\theta(x)$, ancestral sampling is a suitable and effective technique to generate a diverse set of captions.

In practice, the conditional probability distributions $p_\theta(x_t|x_{<t}, I)$ %, often also referred to as the decoder distributions,
are modeled via recurrent LSTM networks or masked convolutions, where we use $h_{t-1}^D$ to refer to its hidden state which summarizes $x_{<t-1}$, while $x_{t-1}$ directly influences the distribution. However, given the preceding words $x_{<t}$ the conditional distribution $p_\theta(x_t|x_{<t}, I)$ has to cover many suitable options to complete the sentence. While LSTM-networks can model complex distributions, we hypothesize that a latent variable $z_t$ which captures the current intention about how to complete the sentence can significantly reduce the complexity of the conditional.

Therefore, instead of directly modeling $p_\theta(x_t|x_{<t}, I)$ which marginalizes across all possible intentions, we are interested in modeling and sampling from the decoder distribution $p_\theta(x_t|x_{<t},z_{\leq t}, I)$, \ie, a distribution conditioned on the current and all previous  intentions, which are however unobserved.

%\begin{figure}
%\centering
%\includegraphics[width=0.6\linewidth]{fig/generation_network.pdf}
%\caption{Computation graph for the generation network. The hidden states of the intention model LSTM and the decoder LSTM are $h_t^T$ and $h_t^D$ respectively. At a given time step $t$, the latent sample $z_t$ depends on all prior words $x_{<t}$ and all prior latent samples $z_{<t}$. The sample $z_t$ along with all prior words $x_{<t}$ predicts $x_{t}$.}
%\label{fig:inf}
%\end{figure}

\noindent\textbf{During inference}, as illustrated in \figref{fig:gen-inf-time-slice}(a), we ensure effective sampling of an intention $z_t$ by modeling the transition through the tuple of all intentions $z = (z_1, \ldots, z_T)$ again via a product of conditionals  $p_\theta(z_t|z_{<t}, x_{<t}, I)$, \ie,
$$
\text{(intention model)}\quad\quad \prod_{t\in\{1, \ldots, T\}} p_\theta(z_t|z_{<t},x_{<t}, I).
$$
%A sampled tuple of intentions $z = (z_1, \ldots, z_T)$ is then used to obtain an image description $x = (x_1, \ldots, x_T)$ by sampling from $p_\theta(x_t|x_{<t},z_t,I)$.
To obtain a description for a given image, as shown in \figref{fig:gen-inf-time-slice}(a), we alternatingly sample from  $p_\theta(z_t|z_{<t},x_{<t}, I)$, which is sometimes referred to as the `prior,' and  from the decoder $p(x_t|x_{<t},z_{\leq t}, I)$.
Different from classical approaches we employ a parametric `intention model' which decomposes temporally.  Note that the `intention distribution' at time $t$, \ie, $p_\theta(z_t|z_{<t},x_{<t}, I)$ is dependent on $x_{<t}$. This is technically correct due to the temporal decomposition, \ie, the distribution at time $t$ can depend on all previously available data. Similar to the decoder, we use an LSTM net to capture the recurrence of the intention model and refer to its latent state via $h_t^T$. %However, we do note that this wouldn't be correct otherwise and the use of $p_\theta(z|x,I)$ may be confusing as we point out below again.

%\begin{figure}[t]
%\centering
%\includegraphics[width=0.6\linewidth]{fig/inference_network.pdf}
%\caption{Computation graph for the encoder network. The hidden states of the forward LSTM and the backward LSTM are $h_t^F$ and $h_t^B$  respectively. At a given time  $t$, the latent sample $z_t$ depends on the entire caption $x$ through the forward and backward models.  }
%\label{fig:train}
%\vspace{-0.52cm}
%\end{figure}

To model the intentions $z$, during training, as illustrated in \figref{fig:gen-inf-time-slice}(b), we encourage the intention model  %$p_\theta(z)$
to fit the approximate posterior $q_\phi(z|x,I)$ which we model using again a product of conditionals, \ie,
$$
\text{(approx.\ posterior)}\quad q_\phi(z|x,I) = \!\!\prod_{t \in \{1, \ldots, T\}}\!\! q_\phi(z_t|z_{t-1},x,I).
$$
The distribution $q_\phi(z|x,I)$ is commonly referred to as the encoder. To adequately capture the intention on how to complete the sentence, as illustrated in \figref{fig:gen-inf-time-slice}(b), we develop a two-stage encoder consisting of a forward stage to model language and a backward stage to summarize intention, \ie, the future of the sentence. We discuss details in \secref{sec:encoder}. %As illustrated in \figref{fig:train} we develop a two-stage encoder

\noindent\textbf{During training} we are given a dataset $\cD = \{(I,x)\}$ consisting of pairs $(I,x)$, each containing an image $I$ and a corresponding caption $x$. We maximize the  data log-likelihood $\sum_{(I,x)\in\cD}\ln p_\theta(x|I)$. For readability we drop the summation over samples in the dataset subsequently. By using the aforementioned decompositions we note that the data log-likelihood $\ln p_\theta(x|I)$ is obtained by marginalizing over the space of intentions, \ie,
\beas
\ln p_\theta(x|I) &\hspace{-0.3cm}=&\hspace{-0.3cm} \ln\sum_{z}  p_\theta(x,z|I)\\ %p_\theta(x|z,I)p_\theta(z|x, I)\\
& \hspace{-0.3cm}=& \hspace{-0.3cm}\ln\sum_{z}\prod_t  \underbrace{\overbrace{p_\theta(x_t|x_{<t},z_{\leq t},I)}^{\text{decoder}} \overbrace{p_\theta(z_t|z_{<t}, x_{<t},I)}^{\text{intention}}}_{p_\theta(x_t, z_t|x_{<t},z_{<t}, I)}.
\eeas
%Note that use of a data-dependent `prior'  makes the first line confusing at the very least. However, checking the temporal decomposition given in the second line more carefully reveals correct modeling of the joint distribution $p(x,z|I)$.

Marginalization over the space of intentions makes maximization of this objective computationally expensive. It is therefore common to utilize an approximate posterior and apply Jensen's inequality which gives the lower bound
$$
\ln p_\theta(x|I) \geq \mathbb{E}_{z \sim q_{\phi}(z|x, I)}\left[\ln \frac{p_\theta(x,z|I)}{q_\phi(z|x,I)}\right].
$$
Combined with the employed temporal decomposition, this yields  the  objective
\bea
\mathbb{E}_{z \sim q_{\phi}(z|x,I)}&&\hspace{-0.8cm}\left[\!\sum_t\left(\ln p_\theta(x_t|x_{<t},z_{\leq t}, I) \right.\right.\label{eq:trainobj}\\
&& \hspace{-0.9cm}\left.\left.+\!\ln p_\theta(z_t|z_{<t},x_{<t},I) \!-\! \ln q_\phi(z_t|z_{t-1},x,I)\right)\right],\nonumber
\eea
which we maximize \wrt parameters $\theta$ and $\phi$. In the following we discuss decoder, prior (intention model) and encoder in more detail. Notice, all their parameters are subsumed in the vectors $\theta$ and $\phi$ and jointly trained end-to-end. A single timestep of all three recurrent models is illustrated in \figref{fig:gen-inf-time-slice}(c).

\vspace{\subsectionReduceTop}
\subsection{Decoder}
\vspace{\subsectionReduceBot}
\label{sec:decoder}
As illustrated in the top part of \figref{fig:gen-inf-time-slice}(c), the decoder is a classical LSTM net. At time $t$ the decoder  yields a multinomial probability distribution $p_\theta(x_t|x_{<t}, z_{\leq t}, I)$ defined over words $x_t\in\cX$.
% {\color{blue}something is wrong in \figref{fig:model}}

While representations of $z_t$ and $x_{t-1}$ are concatenated before being provided as input to the LSTM net, we encode dependence on $x_{<t-1}$ and $z_{<t}$ via its hidden representation $h_{t-1}^D$. Dependence on the image is encoded into the LSTM net via an image embedding obtained from the fc7 layer of a VGG16 network~\cite{simonyan2014very}, pre-trained on the ImageNet dataset. The image embedding is fed as input at every time step of the LSTM,
concatenated with the input word embedding and the sampled vector from the
latent space. %The latent vector dimension can be chosen as 512, 256 or 128.
For all our experiments, we found a 512-dimensional latent vector %the size of the latent vector to be 512
to give the best results (\tabref{tab:ablation}).
%{\color{blue}how is this image embedding used, where is it fed as input?}
%{\color{blue}xxx; talk a little more about the representations as well}

%\input{tables/mrnn_test.tex}
% \input{tables/karparthy_test.tex}
%\begin{figure}[t]
%	\centering
%
%	\includegraphics[width=0.85\linewidth]{fig/model_time_slice.pdf}
%	\caption{Our proposed \seqcvae{} training architecture, shown for a single time slice $t$.  Our model includes three components during training: (1) Two-stage encoder; (2) Intention Model; and (3) Decoder. Details for each of the  components are provided in \secref{sec:approach}. At test time only decoder and intention model are used.}
%	\label{fig:model}
%\end{figure}
\vspace{\subsectionReduceTop}
\subsection{Intention Model}
\label{sec:intention_model}
\vspace{\subsectionReduceBot}

Similar to the decoder we model the intention transition model  $p_\theta(z_t|z_{<t},x_{<t},I)$ as an LSTM net. However, different from the decoder, given $z_{<t}$ and $x_{<t}$, we model $p_\theta(z_t|z_{<t},x_{<t},I)$ as a Gaussian distribution with time-dependent mean $\mu_t^T(z_{<t},x_{<t},I)$ and standard deviation $\sigma_t^T(z_{<t},x_{<t},I)$ obtained from an LSTM net.
The LSTM net input $z_{t-1}^T$ and $x_{t-1}^T$ directly influence $\mu_t^T$ and $\sigma_t^T$. Dependence on $z_{<t-1}$ and $x_{<t-1}$ is encoded via the hidden representation $h_{t-1}^T$. Dependence on the image is encoded into the LSTM net via an image embedding obtained from the fc7 layer of a VGG16 network~\cite{simonyan2014very}, pre-trained on the ImageNet dataset. The 512 dimensional image embedding is fed as input at every time step of the intention model LSTM, concatenated with the output from the previous time step and the word embedding of the previous word. The image embedding, the word embedding  and the latent vector are all 512-dimensional.

\input{tables/mrnn_test.tex}

% {\color{blue}xxx; talk a little more about the representations as well}

During inference, at time $t$ we use a sample $z_t$ from the modeled Gaussian with mean $\mu_t^T(z_{<t},x_{<t},I)$ and standard deviation $\sigma_t^T(z_{<t},x_{<t},I)$ as input for the decoder. However, during training, as illustrated in \figref{fig:gen-inf-time-slice}(c), we use a sample from the encoder. This discrepancy is justified by the fact that part of the training objective given in \equref{eq:trainobj} maximizes the negative KL-divergence
$$
\!\sum_t -\mathbb{E}_{z_t \sim q_\phi(z_t|z_{t-1},x,I)}\left[\ln \frac{q_\phi(z_t|z_{t-1},x,I)}{p_\theta(z_t|z_{<t},x_{<t},I)}\right]
$$
between the intention model and the encoder at each time-step. This is highlighted in \figref{fig:gen-inf-time-slice}(c). Therefore, upon training we want those distributions to be adequately close. This ensures that  samples used during testing are suitable.

More importantly however, note that the encoder $q_\phi(z_t|z_{t-1},x,I)$ depends on the entire sentence $x$ while the intention model $p_\theta(z_t|z_{<t},x_{<t},I)$ only depends on the past observations $x_{<t}$. Consequently, if we construct an adequate encoder and if $p_\theta(z_t|z_{<t},x_{<t},I)$ is accurate, we are able to capture
the intention about how to complete the sentence using   samples from the intention model. We discuss an encoder structure that yielded encouraging results next.

\input{tables/diversity.tex}

\begin{figure*}[t]
	\centering
	 \setlength\tabcolsep{0pt}
	 \renewcommand{\arraystretch}{-2}
	\begin{tabular}{ccc}
\includegraphics[width=0.333\linewidth]{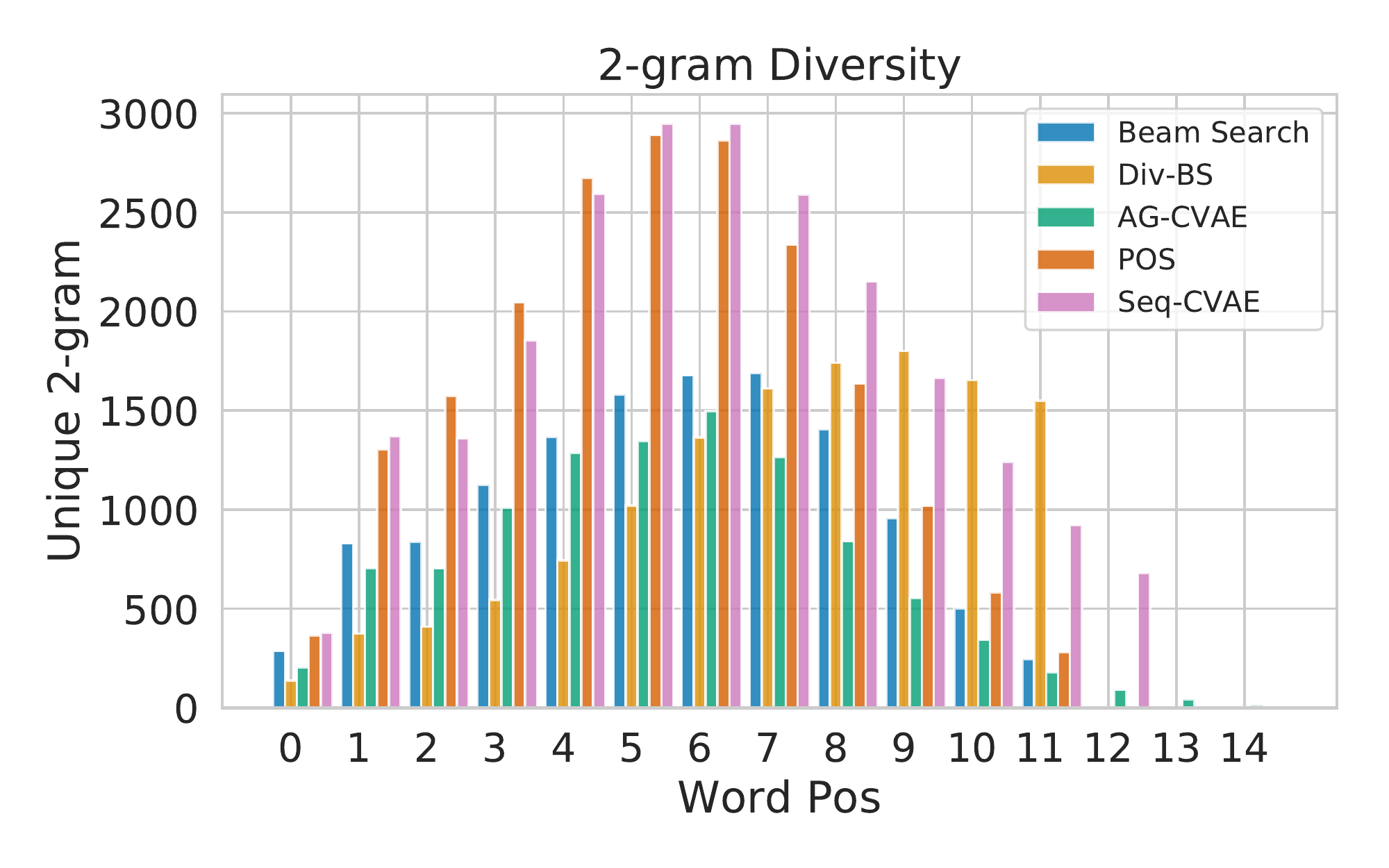}&
\includegraphics[width=0.333\linewidth]{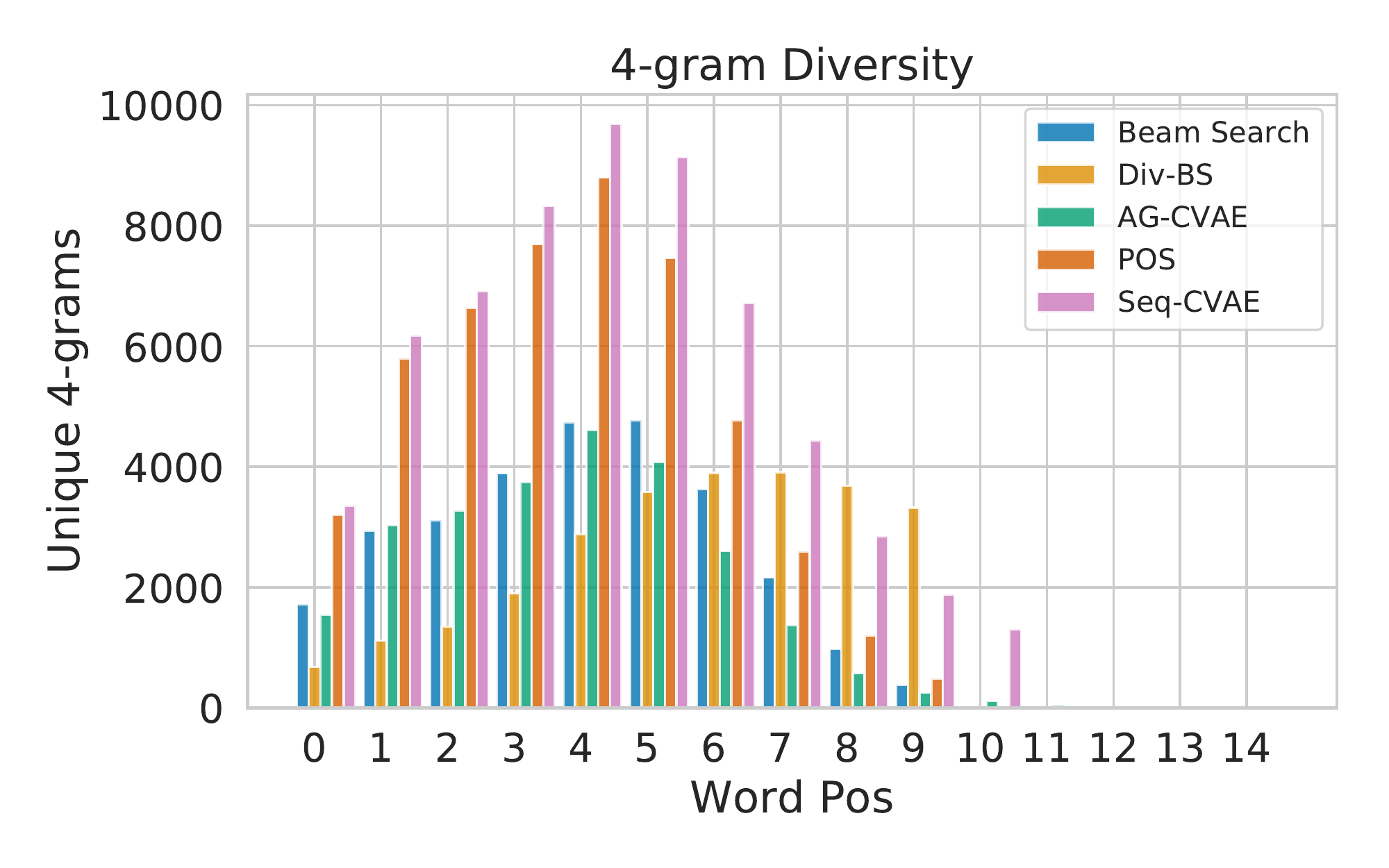}&
\includegraphics[width=0.333\linewidth]{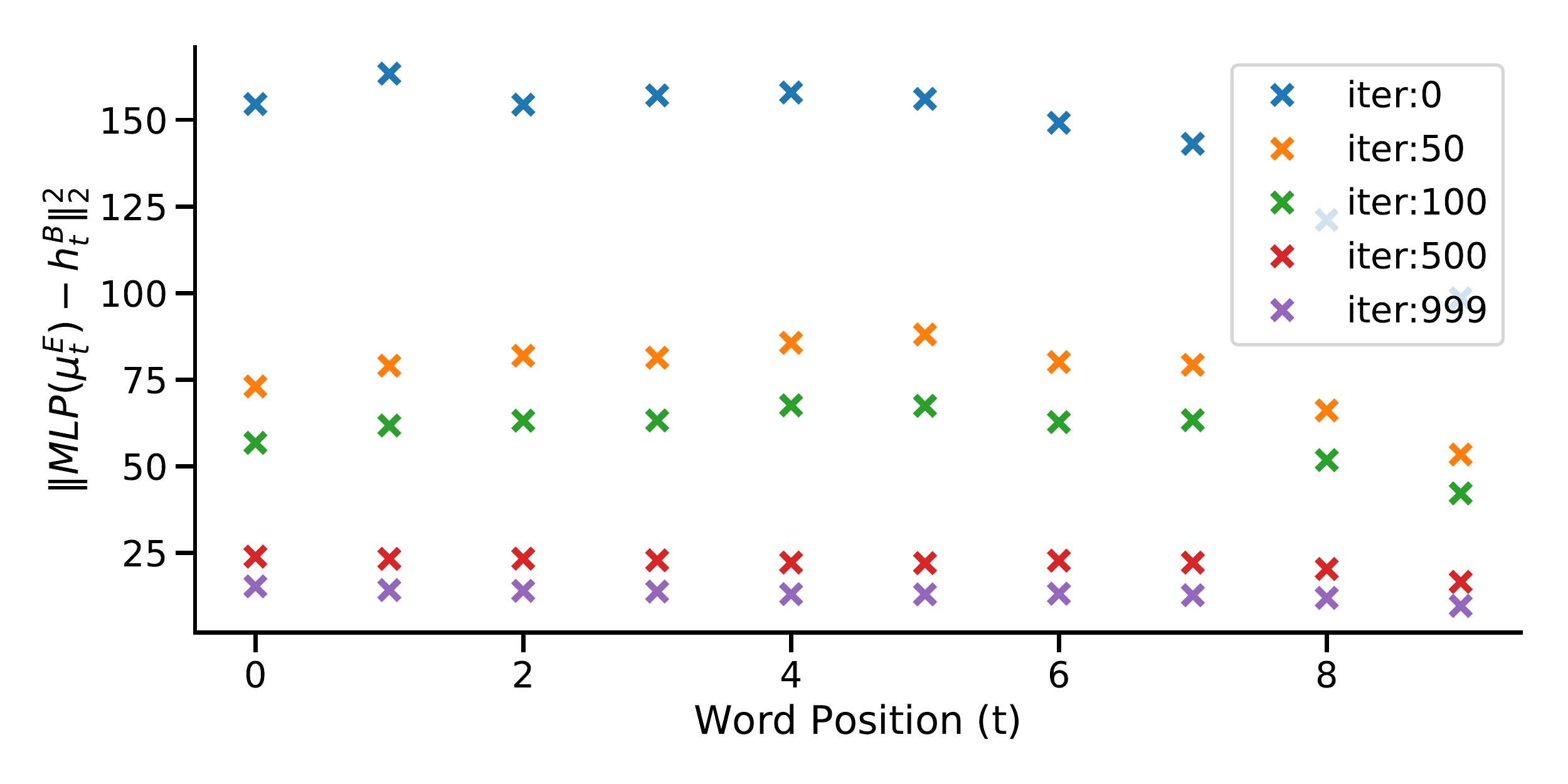}\\
	{\scriptsize (a)}&	{\scriptsize (b)}&	{\scriptsize (c)}
	\vspace{-2pt}
	\end{tabular}
	\caption{
	(a \& b): n-gram diversity across word positions. \seqcvae{}  improves significantly upon many baselines.
	(c): $L_{2}$ distance between the \elmo{} hidden state and the representation inferred by passing the encoder mean $\mu_t^E$ through an MLP. The latter matches $h_t^B$ better as the training progresses. This indicates that the latent space learned by the encoder at a given time $t$ is trained to better regress to word representations which summarize future words.
	}
	\label{fig:ngram-hist}
	\vspace{-0.3cm}
\end{figure*}

\vspace{\subsectionReduceTop}
\subsection{Encoder to Expose Intention}
\label{sec:encoder}
\vspace{\subsectionReduceBot}
To adequately encode the intention, \ie, the future of a sentence, we need to construct a model which contains at time $t$ information about the entire sentence rather than only its past. To achieve this, we develop a two-stage encoder $q_\phi(z_t|z_{t-1},x,I)$ which models at time $t$ a Gaussian distribution with mean $\mu_t^E(z_{t-1},x,I)$ and standard deviation $\sigma_t^E(z_{t-1},x,I)$ modulated by multiplying with the exponentiated function value $F_\phi(\mu_t^E,x,I)\in\mathbb{R}$, \ie,
$$
q_\phi(z_t|z_{t-1},x,I)\; \propto\; \cN(z_t | \mu_t^E, \sigma_t^E) \cdot \exp F_\phi(\mu_t^E,x,I).
$$
Note that multiplication with the exponentiated function value $F_\phi(\mu_t^E, x, I)$ doesn't change the fact that $q_\phi$ is a valid distribution over the latent space. Importantly however, multiplication permits to add  the term $-F_\phi(\mu_t^E,x,I)$ to the objective given in \equref{eq:trainobj}. As detailed below, we will use $F_\phi$ to encourage $\mu_t^E$ to better capture the future of a sentence.

The first stage of the encoder captures the past via a classical forward LSTM net with hidden states referred to as $h_t^F$. The second stage of the encoder captures the future via a backward LSTM net with hidden states referred to as $h_t^B$. We subsequently combine both via a multi-layer perceptron (MLP) net which yields mean $\mu_t^E(z_{t-1},x,I)$ and standard deviation $\sigma_t^E(z_{t-1},x,I)$ of the Gaussian  distribution.

To ensure that the mean $\mu_t^E(z_{t-1},x,I)$ more closely resembles the information obtained from the backward pass we choose
\be
F(\mu_t^E,x,I) = \lambda\|g(\mu_t^E(z_{t-1},x,I)) - h_t^B\|_2^2,
\label{eq:Ffunc}
\ee
where $g$ is another MLP which maps  $\mu_{t}^{E}$ to fit $h_{t}^{B}$. The latter is 512-dimensional in our case. $\lambda$ is a hyper-parameter set to $5e^{-4}$. For the backward LSTM we use the pre-trained \elmo{}~\cite{Peters:2018} model, with a hidden dimension of 512. \elmo is a deep bidirectional language model trained on 1 Billion Word Language Model Benchmark~\cite{chelba2013one}. We only use the backward part of the model. Word representations taken from this backward pass at any time $t$ are a good encoding of the future $x_{>t}$. \elmo is not fine-tuned through training.

 %\elmo because {\color{blue}xxx}.

%{\color{blue}how are image embeddings used here?}

% {\color{blue}describe the representations and MLPs; also discuss ELMO}

% \begin{figure}
% \centering
% \includegraphics[width=0.5\linewidth]{fig/kaw.png}
% \caption{Encoder}
% \label{fig:encoder}
% \end{figure}

%% file: tables/mrnn_test.tex
\begin{table}[t!]
%\vspace{-.5cm}
\centering
\scriptsize

\setlength\tabcolsep{2pt}
\begin{tabular}{clllllllll}
\toprule
&	Method  & \multicolumn{8}{c}{Best-1 Oracle Accuracy on M-RNN Test Split} \\
&	& B4 & B3 & B2 & B1 & C &  R & M & S \\
\midrule

\multirow{5}{*}{\rotatebox[origin=c]{90}{\specialcell{Beam size \\\#samples: 20}}} &	Beam search &  \bf 0.489 &  \bf 0.626 & \bf 0.752 & \bf 0.875 & \bf 1.595 & \bf 0.698 & \bf 0.402 & \bf 0.284\\
	&Div-BS~\cite{Vijayakumar2018DiverseBS} &  0.383 & 0.538 & 0.687 & 0.837 & 1.405 & 0.653  & 0.357  & 0.269 \\
	&AG-CVAE~\cite{Wang2017DiverseAA} &  0.471 & 0.573 & 0.698 & 0.834 & 1.259 & 0.638 & 0.309 & 0.244\\
	&POS~\cite{deshpande2019fast} &  0.449 & 0.593 & 0.737 & 0.874 & 1.468 & 0.678 & 0.365 & 0.277\\
    &\specialcell{Seq-CVAE}  &  0.445 & 0.591 & 0.727 & 0.870 & 1.448 & 0.671 & 0.356 & 0.279\\

\midrule
\multirow{5}{*}{\rotatebox[origin=c]{90}{\specialcell{Beam size \\\#samples: 100}}} &	Beam Search &  \bf 0.641 & \bf 0.742 & \bf 0.835 & \bf 0.931 & \bf 1.904 & \bf 0.772 & \bf 0.482 & \bf 0.332\\
	&Div-BS~\cite{Vijayakumar2018DiverseBS}  &  0.402  & 0.555 & 0.698 & 0.846  &  1.448 & 0.666 & 0.372  & 0.290\\
	&AG-CVAE~\cite{Wang2017DiverseAA} &  0.557 & 0.654 & 0.767 & 0.883 & 1.517 & 0.690 & 0.345 & 0.277\\
	&POS~\cite{deshpande2019fast} &  0.578 & 0.689 & 0.802 & 0.921 & 1.710 & 0.739 & 0.423 & 0.322\\
    &\specialcell{Seq-CVAE}  &  0.575 & 0.691 & 0.803 & 0.922 & 1.695 & 0.733 & 0.410 & 0.320 \\
\bottomrule
\end{tabular}%
%\vspace{-.1cm}
\caption{{\bf Best-1-Oracle Accuracy.} Our \seqcvae{} method obtains high scores on standard captioning metrics. We obtain an accuracy comparable   to both the very recently proposed POS approach~\cite{deshpande2019fast} which uses a part-of-speech prior and also to the AG-CVAE method~\cite{Wang2017DiverseAA}. Both these methods  use additional information in the form of object vectors from a Faster-RCNN~\cite{ren2015faster} during inference. In contrast, we do not use any additional information during inference. To calculate the best-1-accuracy, we report the caption with highest CIDEr score %as an oracle,
from all the sampled captions ($\#$ samples = 20 or 100). Beam search, although obtaining the highest CIDEr score, is known to be extremely slow and significantly less diverse.
}
\vspace{-.6cm}
\label{tab:mrnn_split}
\end{table}

%% file: tables/diversity.tex
\begin{table}[!t]
%\vspace{-0.5cm}
\scriptsize
\centering
\begin{center}
\setlength{\tabcolsep}{4pt}
%\renewcommand{\arraystretch}{1.2}
%\begin{tabularx}{0.9\textwidth}{1Y1Y1Y1Y1Y11Y1}
\begin{tabular}{ccccccc}
\toprule
	&Method & \multirow{2}{*}{\specialcell{Distinct \\ Captions}} & \multirow{2}{*}{\specialcell{\# Novel \\Sentences}} & \multirow{2}{*}{\specialcell{mBleu-4}} & \multicolumn{2}{c}{$n$-gram Diversity}\\

	\cmidrule{6-7}
	&&   &  &   & Div-1 & Div-2 \\
\midrule
	\multirow{5}{*}{\rotatebox[origin=c]{90}{\specialcell{Beam size \\\#samples: 20}}}&Beam search  &  {\bf 100\%} &  2317 & 0.77 & 0.21 & 0.29\\
	&Div-BS~\cite{Vijayakumar2018DiverseBS}  &  {\bf 100\%} &  3106 & 0.81 & 0.20 & 0.26\\
	&AG-CVAE~\cite{Wang2017DiverseAA}   & 69.8\% & 3189 & 0.66  & 0.24 & 0.34\\
	&POS~\cite{deshpande2019fast}  & 96.3\% & 3394 & 0.64 &  0.24 & 0.35\\
	&\specialcell{Seq-CVAE}  & 94.0\% & {\bf 4266} & {\bf 0.52} & {\bf 0.25} & {\bf 0.54}\\
\midrule
	\multirow{5}{*}{\rotatebox[origin=c]{90}{\specialcell{Beam size \\\#samples: 100}}}&Beam search  & {\bf 100\%} & 2299 & 0.78 & 0.21 & 0.28\\
	&Div-BS~\cite{Vijayakumar2018DiverseBS} & {\bf 100\%} & 3421  & 0.82 & 0.20 & 0.25\\
	&AG-CVAE~\cite{Wang2017DiverseAA} & 47.4\% & 3069 & 0.70 &  0.23 & 0.32\\
	&POS~\cite{deshpande2019fast} & 91.5\% &  3446 &  0.67 & 0.23 & 0.33\\
	&\specialcell{Seq-CVAE} & 84.2\% & {\bf 4215} & {\bf 0.64} & {\bf 0.33} & {\bf 0.48}\\
	\specialrule{.2em}{.1em}{.1em}
	5&Human  & 99.8\% & - &  0.51 &  0.34 &  0.48 \\
\bottomrule
\end{tabular}%
%\vspace{-0.1cm}
\caption{{\bf Diversity Statistics.} We report the  number of novel sentences (sentences never seen during training)  for each method. Beam search and diverse beam search (Div-BS) produce the least number of novel sentences. POS~\cite{deshpande2019fast} uses additional information in the form of part-of-speech tokens and object detections from Faster-RCNN~\cite{ren2015faster}. AG-CVAE~\cite{Wang2017DiverseAA} also uses additional information in the form of object vectors. Our \seqcvae{} with \elmo{} doesn't use any additional information during inference and produces 4278/5000 novel sentences. Our method also yields  significant improvements on  2-gram diversity, producing $\approx20\%$ more unique 2-grams for 20 samples and a $\approx15\%$ improvement for 100 samples when compared to the runner-up, \ie, POS~\cite{deshpande2019fast}. The model also provides the lowest m-Bleu-4, which shows that for each image the diverse captions  are most different from each other. Beam search has the highest m-Bleu-4, which shows that  all the distinct captions don't differ  from each other at many word locations.}
\label{tab:diversity}
\end{center}
\vspace{-0.8cm}
\end{table}

%% file: sections/results.tex
% !TEX root = ../main.tex

%\input{tables/ablation.tex}
\vspace{-0.3cm}
\vspace{\sectionReduceTop}
\section{Results}
\vspace{\sectionReduceBot}
\label{sec:exp}
\vspace{-0.2cm}
In the following, we first describe the dataset along with the competitive baselines for diverse captioning and the evaluation metrics used. We then present our results.

\noindent
\textbf{Dataset:} We use the challenging \textbf{MSCOCO} dataset~\cite{lin2014microsoft} for our experiments. Following the approach in~\cite{deshpande2019fast, Wang2017DiverseAA}, we perform our analysis on %two train/val/test splits: (1) the split by Karpathy \etal~\cite{KarpathyCVPR2015} which has 113,287 train, 5,000 val and 5,000 test images; and (2)
the split of M-RNN~\cite{MaoARXIV2014} which has 118,287 train, 4,000 val and 1,000 test images. Additional results are deferred to the supplementary material.

\noindent
{\bf Methods:} We refer to our proposed approach as \seqcvae{}. We also provide results of our proposed approach without the \elmo based backward LSTM and without the data-dependent intention model.

We compare the results to the diverse captioning approach of~\cite{deshpande2019fast} which uses part-of-speech as a prior and refer to the method via {\bf POS}. We also compare to the additive Gaussian conditional VAE-based  diverse captioning
method of Wang \etal~\cite{Wang2017DiverseAA}, denoted by {\bf AG-CVAE} which uses object detections as additional information. Moreover, we compare to beam search denoted as {\bf Beam search} and diverse beam search~\cite{Vijayakumar2018DiverseBS} referred to as {\bf Div-BS} applied to standard captioning methods based on convolutions~\cite{AnejaConvImgCap17} and LSTM networks~\cite{karpathy2015deep}.

\noindent
{\bf Evaluation criteria:} We compare all aforementioned methods via the following accuracy and diversity metrics:
\begin{itemize}[itemsep=-1ex,topsep=0pt,partopsep=0pt]
% \item {\bf Accuracy.} There are two meaningful ways of measuring accuracy in context of diversity.
% Top-1-Accuracy, and Top-n-Accuracy, evaluated on standard image captioning metrics. We compare Top-1T-Accuracy in \secref{sec:top1} and Top-n-Accuracy in \secref{sec:topn}.
\item {\bf Accuracy.} In  \secref{sec:top1}, we report the Top-1-Accuracy evaluated  on the standard image captioning metrics (Bleu-1 to Bleu-4, CIDEr, ROUGE, METEOR and SPICE, each abbreviated with its initial).
\item {\bf Diversity.} Our diversity evaluation is presented in \secref{sec:diversity}
\end{itemize}

\vspace{\subsectionReduceTop}
\subsection{\bf Top-1-Accuracy}
\vspace{\subsectionReduceBot}
\label{sec:top1}
We use  CIDEr as an oracle to pick the top-1 caption from a set of generated diverse captions for a given image.

%\noindent
%{\bf Oracle}
Following the approach of Deshpande \etal~\cite{deshpande2019fast} and Wang \etal~\cite{Wang2017DiverseAA}, the top-1 caption is chosen as the caption with the maximum score calculated with the ground-truth test captions as references.
The oracle metric provides the maximally possible top-1 accuracy that a given model can achieve. In \tabref{tab:mrnn_split} %and  \tabref{tab:karparthy_split}
we show that our \seqcvae performs on par with the best baselines on the M-RNN split~\cite{MaoARXIV2014}. % and Karpathy split~\cite{KarpathyCVPR2015}.

Specifically, in \tabref{tab:mrnn_split} %and  \tabref{tab:karparthy_split}
%<<<<<<< HEAD
we show for 20 and 100 samples, that the proposed approach obtains a CIDEr score of 1.448 and 1.695 respectively, on par with POS. Importantly, we emphasize that the proposed approach doesn't use any additional information in the form of part-of speech tags or object vectors from a Faster-RCNN~\cite{ren2015faster} during inference. Note that all the other scores, are comparable to the POS~\cite{deshpande2019fast} approach while improving upon the   AG-CVAE~\cite{Wang2017DiverseAA} method. The latter is the only other VAE based method which exhibits stochasticity when producing diverse captions. Note that although beam search obtains the best scores, it is known to be  slow and less diverse as shown in \secref{sec:diversity}. %, least diverse.
%=======
%we show for 20 and 100 samples, that the proposed approach obtains a CIDEr of 1.448 and 1.695 respectively, on par with POS. Importantly, we emphasize that the proposed approach doesn't use any additional information in the form of part-of speech tags or object vectors from a Faster-RCNN~\cite{ren2015faster} during inference. Note that all the other scores, are comparable to the POS~\cite{deshpande2019fast} approach while improving upon the   AG-CVAE~\cite{Wang2017DiverseAA} method, which is the only other VAE based method which exhibits stochasticity when producing diverse captions. Note that although beam search obtains the best scores, it is known to be  slow and less diverse as shown in \secref{sec:diversity}. %, least diverse.
%>>>>>>> d1b98c3f622d209732aa5380f904d91ace3955d6
%{\color{blue}xxx}.

% \begin{figure}[t]
% 	%\vspace{-0.2cm}
% 	\centering
% 	\begin{subfigure}{.5\linewidth}
% 		\centering
% 		\includegraphics[width=.7\linewidth]{fig/T_SNE_w_ELMO.png}
% 		\caption{}

% 	\end{subfigure}%
% 	\begin{subfigure}{.5\linewidth}
% 		\centering
% 		\includegraphics[width=.7\linewidth]{fig/T_SNE_wo_ELMO.png}
% 		\caption{}

% 	\end{subfigure}
% 	\vspace{\captionReduceTop}
% 	\vspace{-0.1cm}
% 	\caption{\tsne{} plots of the means  $\mu_t^T$ obtained from the intention model, learned {\bf {with \elmo{} (a)}} and {\bf{without \elmo{} (b)}}. Notice that using the \elmo{} representation, the model learns to better disentangle the means per word.}
% 	\label{fig:tsne_word_pos}
% 	\vspace{-0.2cm}
% \end{figure}
\vspace{\subsectionReduceTop}
\subsection{\bf Diversity Evaluation}
\label{sec:diversity}
\vspace{\subsectionReduceBot}
To ensure comparability to the baselines, our diversity numbers are calculated on the M-RNN split~\cite{MaoARXIV2014}. In \tabref{tab:diversity} we show the diversity results using the following metrics:

\noindent
{\bf (1) Uniqueness.} The number of distinct captions generated by sampling from the latent space. We show that the proposed method produces 18.48/20 (92.4\%) and 80.9/100 (80.9\%) unique sentences. %Sampling from the latent space introduces randomness whereas
Note that beam search and Div-BS are deterministic and are guaranteed to generate 100\% unique captions.
Similarly, POS is completely deterministic and ensures a large number of unique captions via a strong connection between generated words and a hard-coded `latent space' which depends on part-of-speech tags and is learned in a fully supervised manner.
In contrast, AG-CVAE, just like the proposed approach, has a flexible latent space. Compared to AG-CVAE, the proposed approach generates significantly more distinct captions. %which is a VAE based approach and samples considerably less unique  captions.

\noindent
{\bf (2) Novel Sentences.} Novel sentences are those sentences which were never observed in the training data. We see that \seqcvae produces significantly more  novel sentences than any other baseline. This is remarkable and illustrates the ability to emit novel words that form reasonable sentences, particularly when considering that accuracy metrics are on par with the best performing baselines.
In \tabref{tab:diversity}, we show that our approach produces >4000 novel captions among the 5000 captions chosen. We choose the top-5 generated captions per image, ranked by CIDEr, using consensus re-ranking following the approach in~\cite{Devlin2015ExploringNN,Wang2017DiverseAA}.

\input{tables/diversity_ablation.tex}

\noindent
{\bf (3) Mutual Overlap -- (mBleu-4).} m-Bleu-4 measures the difference between predicted diverse captions. Specifically, for a given image, we calculate the Bleu-4 metric for every one of the $K$ diverse captions \wrt the remaining $K-1$ and average across all  test images. A lower value of m-Bleu indicates more diversity. Again we observe that the proposed approach  significantly improves upon all  baselines. As before, we use top-5 generated captions, ranked by CIDEr, using consensus re-ranking~\cite{Devlin2015ExploringNN,Wang2017DiverseAA}.

\noindent
{\bf (4) n-gram diversity -- (Div-$n$):} For Div-$n$ scores, we measure the ratio of distinct $n$-grams to the total number of words generated per set of diverse captions. Higher is better. Again we observe that our approach significantly improves upon the baselines, particularly when considering 2-grams. For instance, we improve from 0.35 to 0.54 when considering 20 samples and from 0.33 to 0.48 when considering 100 samples. This is encouraging because it  illustrates the ability of our approach to produce fitting yet diverse descriptions without using any additional information.

\noindent
%<<<<<<< HEAD
{\bf (5) Unique n-grams.} We measure the unique 2-grams and the unique 4-grams produced by our model in  \figref{fig:ngram-hist}(a,~b). We observe that our model produces the largest number of 4-grams for all word positions until position 8. We produce a comparable number of unique 2-grams as POS~\cite{deshpande2019fast}. To compute the numbers we use 20 samples from the latent space for each of the 1000 test images. %, and report the numbers for each method.
%=======
%{\bf (5) Unique n-grams.} We measure the unique 2-grams and the unique 4-grams produced by our model in  \figref{fig:ngram-hist}~(a,~b). We observe that our model produces the largest number of 4-grams for all word positions until position 8. We produce a comparable number of unique 2-grams as POS~\cite{deshpande2019fast}. To compute the numbers we use 20 samples from the latent space for each of the 1000 test images. %, and report the numbers for each method.
%>>>>>>> d1b98c3f622d209732aa5380f904d91ace3955d6
This higher number of unique 4-grams, indicates that  a model is not just producing unique words, but also unique combinations of words.

\input{tables/ablation.tex}

%\begin{figure*}[t]
%\vspace{-0.4cm}
%	\centering
%	\includegraphics[width=1.\linewidth]{fig/qual_captions_3.pdf}
%	\vspace{-0.5cm}
%	\caption{Qualitative results illustrating captions obtained from different image captioning methods.}
%	\label{fig:qualres}
%	\vspace{-0.3cm}
%\end{figure*}

% \input{tables/ablation.tex}
% \begin{figure}[h]
% %\vspace{-0.2cm}
% \centering
% \includegraphics[width=0.7\linewidth]{fig/WordGroups.pdf}
% \caption{\tsne{} plot of the means  $\mu_t^T$ learned by the intention model, mapped to the words produced by the decoder. Notice that similar words like `suit,' `cheese,' \etc are grouped  in tight clusters.}
% \label{fig:word_groups}
% \vspace{-0.3cm}
% \end{figure}

\begin{figure*}[t]
    \centering
    \includegraphics[width=1.0\linewidth]{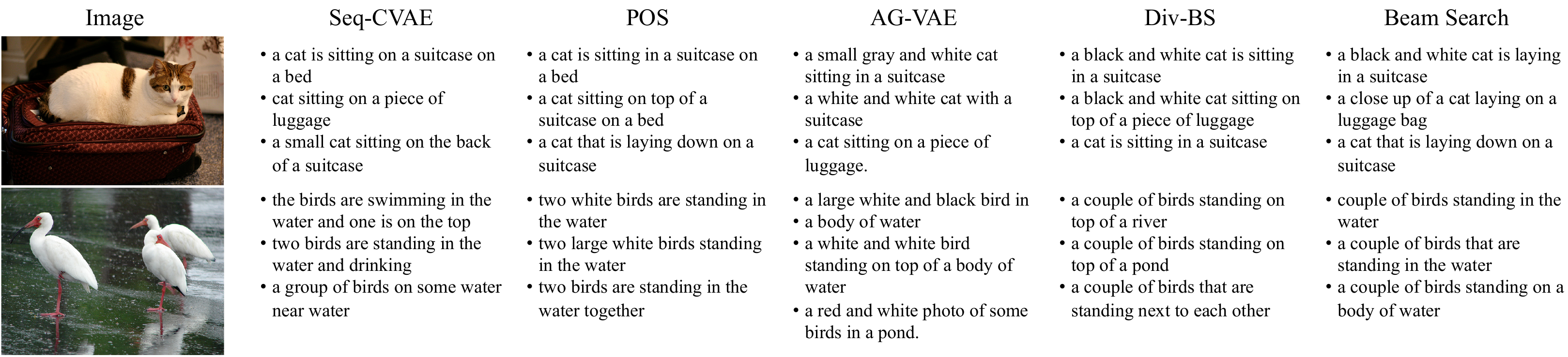}
    \caption{Qualitative results illustrating captions obtained from different image captioning methods.}
    \label{fig:word_groups}
    \vspace{-0.7cm}
\end{figure*}

\begin{figure}[t]
  %\vspace{-0.2cm}
  \centering
  \begin{subfigure}{.5\linewidth}
      \centering
      \includegraphics[width=.7\linewidth]{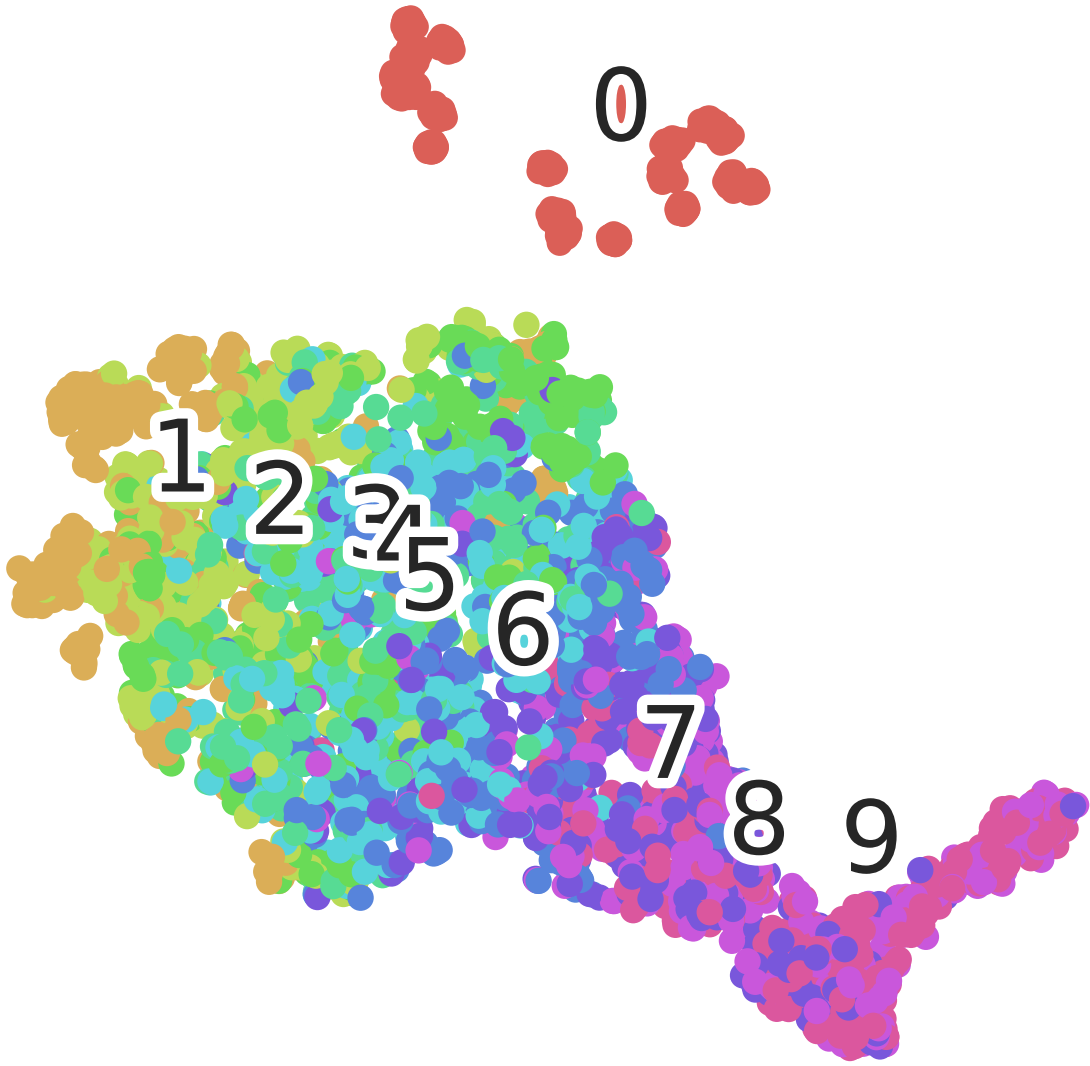}
      \caption{}

  \end{subfigure}%
  \begin{subfigure}{.5\linewidth}
      \centering
      \includegraphics[width=.7\linewidth]{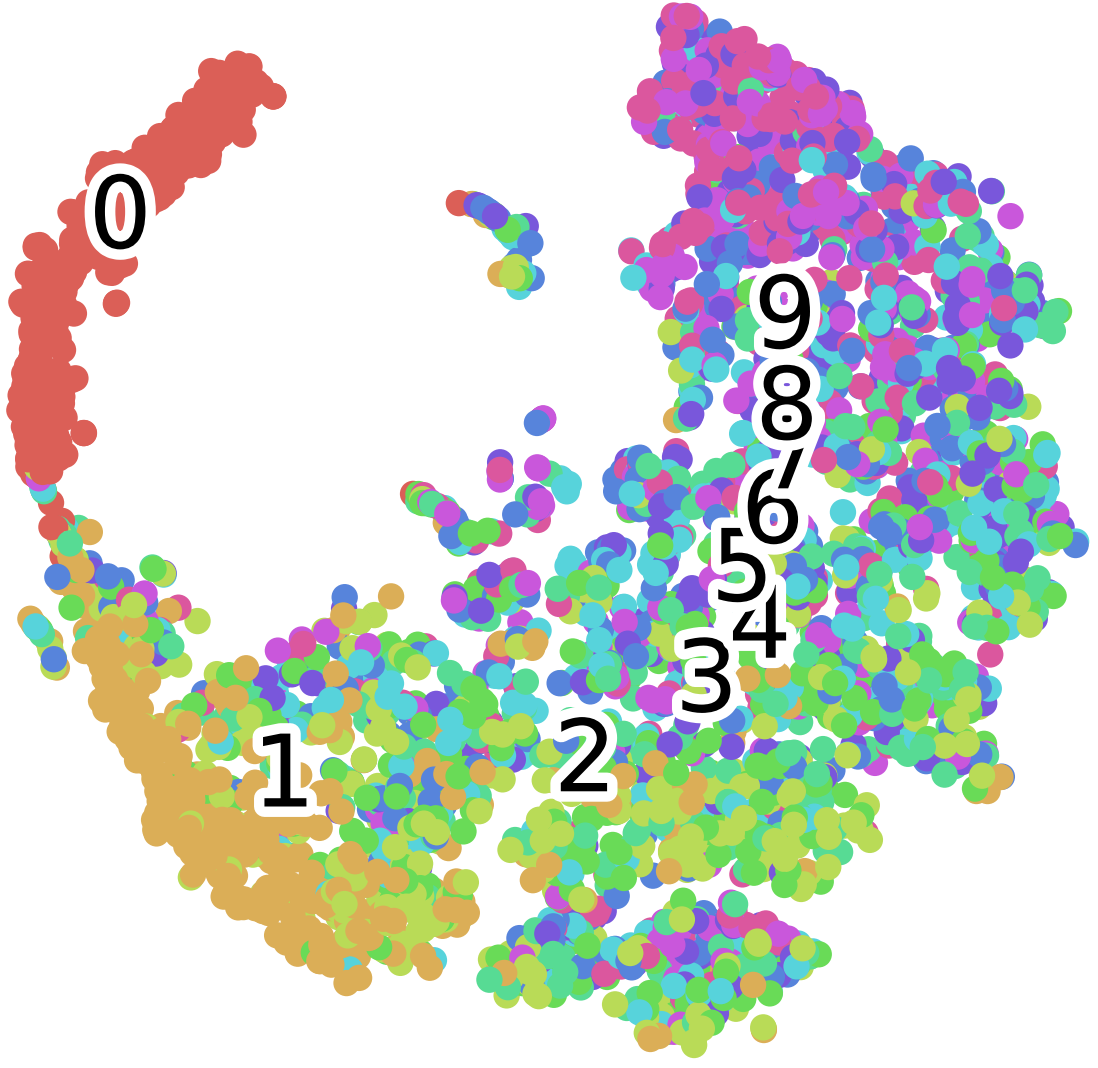}
      \caption{}

  \end{subfigure}
%  \vspace{\captionReduceTop}
  \vspace{-0.0cm}
  \caption{\tsne{} plots of the means  $\mu_t^T$ obtained from the intention model, learned {\bf {with \elmo{} (a)}} and {\bf{without \elmo{} (b)}}. Notice that with \elmo{} representation, the model  better disentangles the means per word.}
  \label{fig:tsne_word_pos}
  \vspace{-0.cm}
\end{figure}

\begin{figure}[t]
\centering
\includegraphics[width=0.6\linewidth]{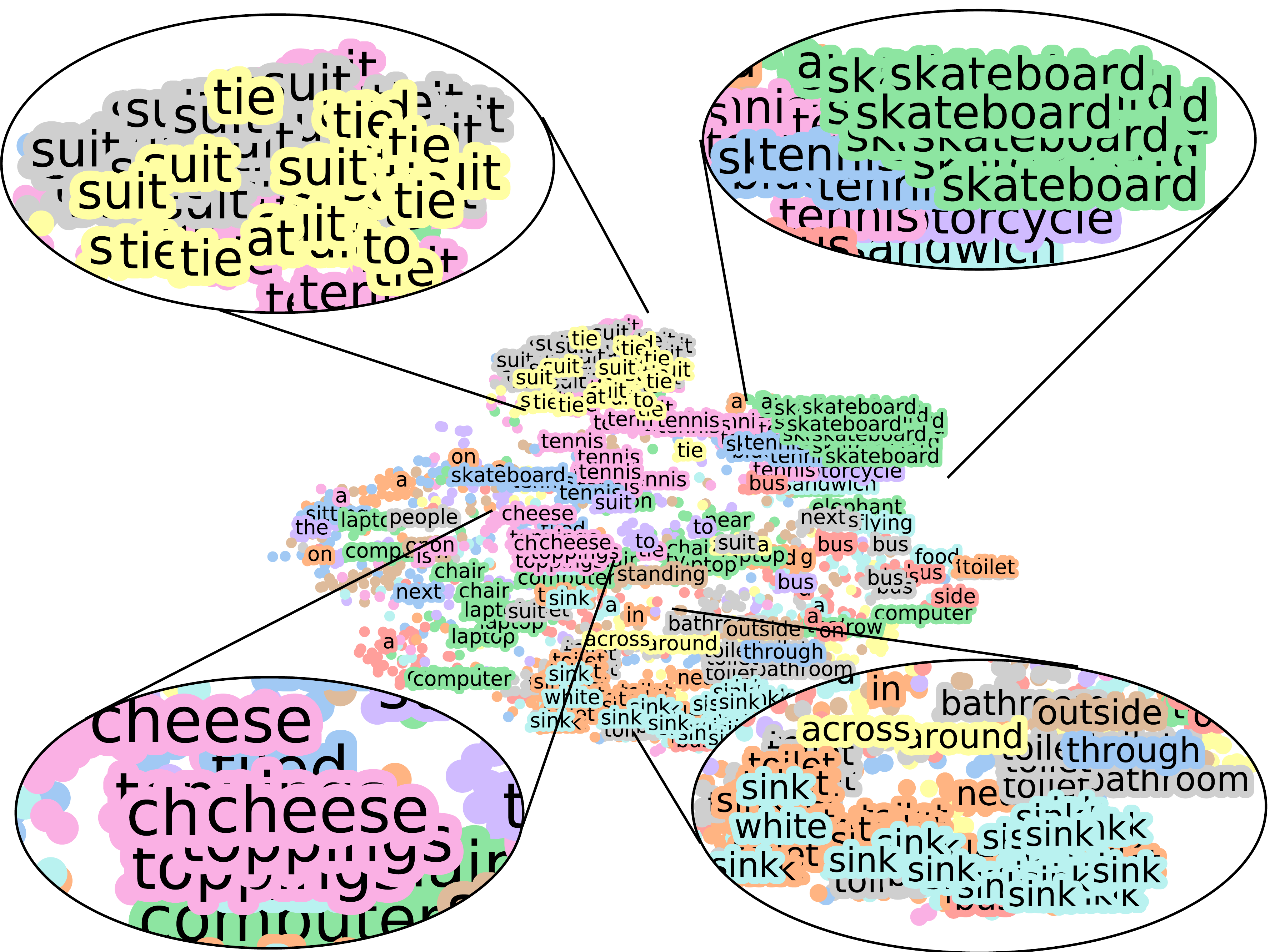}
\caption{\tsne{} plot of the means  $\mu_t^T$ learned by the intention model, mapped to the words produced by the decoder. Notice that similar words like `suit,' `cheese,' \etc are grouped  in tight clusters.}
\label{fig:word_groups}
 \vspace{-0.7cm}
\end{figure}

\vspace{\subsectionReduceTop}
\subsection{Ablation Study}
\vspace{\subsectionReduceBot}

\noindent
{\bf (1) Is \elmo the reason for good performance?}
To analyze if the improvements are only due to a strong language model like \elmo, we replaced  \elmo with a backward RNN trained on MSCOCO training data. The performance in both diversity metrics and CIDEr, are comparable to our \elmo based model, indicating the high performance gains are not just from using a strong pretrained language model (Tab.~\ref{tab:div-acc-abl} row 7, Seq-CVAE(BRNN)).

\noindent
{\bf (2) Using a single latent variable.}  Accuracy and diversity drop when using a single $z$ (both with and without ELMo) to encode the entire sentence (Tab.~\ref{tab:div-acc-abl} rows 3, 4; CVAE), due to  posterior collapse. Also, the latent space differs per word (\figref{fig:tsne_word_pos}, \figref{fig:word_groups}). A single $z$ doesn't efficiently  encode this.

\noindent
{\bf (3) Using a single LSTM for Encoder, Decoder and Transition Network.}
Different distributions  are best served by their own individual
representation. Following the approach in~\cite{goyal2017z} using the same LSTM for all networks leads to inferior results (Tab.~\ref{tab:div-acc-abl} row 2, Z-forcing).

\noindent
{\bf (4)  Using a constant Gaussian distribution per word.} Replacing the LSTM based learnable  intention model with a constant Gaussian reduces performance (both with and without \elmo) indicating the importance to distill intent via the backward LSTM into a sequential
latent space (Tab.~\ref{tab:div-acc-abl} rows 5, 6; Seq-CVAE+$\cN$).

\noindent
{\bf (5) Conditioning over different $z$ and $x$.}
The results are summarized in \tabref{tab:ablation} and averaged over 10 runs. We show results without using the \elmo based backward LSTM in the encoder (see column titled \elmo), without using a data-dependent intention model ($z_t|z_{<t}$, \ie, the intention model isn't conditioned  on $x_{<t}$), and without using any intention (\ie, the intention model is a zero mean unit variance Gaussian $\cN$ for all word positions $t$). %to our method which includes ELMO (as discussed in {\color{blue}xxx}) and to our method which is based on Shared LSTM (as discussed in {\color{blue}xxx}).
%For different latent dimensions (128, 256, and 512), we observe in \tabref{tab:ablation} that  usage of \elmo  results in significant improvements.
Note, based on the CIDEr metric, data dependent intention doesn't contribute much when sampling 20 captions. However, data dependent intention has a slight edge when sampling 100 captions, irrespective of the chosen latent dimension. %This is intuitive because {\color{blue}xxx}.
Note that the standard deviations shown in \tabref{tab:ablation} are fairly small.
\vspace{-0.1cm}

\vspace{\subsectionReduceTop}
\subsection{Latent Space Analysis}
\vspace{\subsectionReduceBot}
To further understand the intricacies of the learned latent space we  analyze its behavior in the following. %In this entire analysis we use the mean $\mu_t^T$ captured by the intention model unless otherwise specified.

%\begin{figure}[t]
%\includegraphics[width=1.0\linewidth]{fig/L2-Dist-10.pdf}
%\caption{$L_{2}$ distance between the \elmo{} hidden state and the representation inferred by passing the encoder mean $\mu_t^E$ through an MLP. The latter matches $h_t^B$ better as the training progresses. This indicates that the latent space learned by the encoder at a given time $t$ is trained to better regress to word representations which summarize future words.}
%\label{fig:elmo_dist}
%\end{figure}

%In \figref{fig:elmo_dist}
In \figref{fig:ngram-hist}(c) we illustrate for different training iterations (see legend) and word positions $t$ the averaged $F(\mu_t^E, x, I)$ given in \equref{eq:Ffunc}, \ie, the L2 distance between the \elmo representation $h_t^B$ obtained via the backward LSTM and the \elmo representation inferred by passing the encoder mean $\mu_t^E$ through the MLP $g$. Intuitively we observe models at later iterations to better match the \elmo representation $h_t^B$. %Very intuitively, we also observe that \elmo representations at later word positions are much easier to capture which is partly due to the fact that they are less diverse.

To further investigate whether the mean $\mu_t^T$ of the intention model used during inference captures meaningful transitions, we illustrate in \figref{fig:tsne_word_pos}(a,~b) \tsne~\cite{maaten2008visualizing} plots of means obtained from different images and colored by word position $t$. We can clearly observe that the word positions of $\mu_t^T$ are much better grouped when using \elmo representation (\figref{fig:tsne_word_pos}~(a)) whereas they are more cluttered when training \seqcvae{} without \elmo representations. We verified this analysis across multiple runs.

In \figref{fig:word_groups} we illustrate a \tsne plot of $\mu_t^T$ based on words emitted by the decoder. We clearly observe clusters of words like `woman,' `man,' `dog,' `horse,' `group,' `bathroom,' `toilet,' \etc. This grouping is encouraging as it illustrates how we can control individual emitted words by transitioning from one representation to another. Results for this transition are illustrated in \figref{fig:teaser}.

%\vspace{-0.25cm}

\input{fig/qualitative_figure.tex}

\vspace{\subsectionReduceTop}
\subsection{Qualitative Results}
\vspace{\subsectionReduceBot}
%We illustrate qualitative results in \figref{fig:qualres}.
We show a transition between two sampled captions in \figref{fig:transition}. We linearly interpolate the latent vectors at all word positions between two sampled descriptions. %We observe the proposed approach to {\color{blue}xxx}.

%% file: tables/diversity_ablation.tex
\begin{table}[t]
%\vspace{-0.2cm}
% \tiny
\centering
% \footnotesize
\scriptsize
\begin{center}
\setlength{\tabcolsep}{0.3pt}
\renewcommand{\arraystretch}{1.0}
%\begin{tabularx}{0.9\textwidth}{1Y1Y1Y1Y1Y11Y1}
\begin{tabular}{ccccccccc}
\toprule
	 \multirow{2}{*}{\specialcell{ Method}} &
	 \multirow{2}{*}{\specialcell{ELMo}}&
	 \multirow{2}{*}{\specialcell{Distinct \\ Captions}} & \multirow{2}{*}{\specialcell{\# Novel \\Sentences}} & \multirow{2}{*}{\specialcell{mBleu-4}} & \multicolumn{2}{c}{$n$-gram Diversity} & \multirow{2}{*}{\specialcell{CIDEr}}\\

	\cmidrule{6-7}
	 &&   & &  & Div-1 & Div-2 &  \\
\midrule
% 	\multirow{5}{*}{\rotatebox[origin=c]{90}{\specialcell{Beam size \\\#samples: 20}}}&POS~\cite{deshpande2019fast}  & {\bf96.3\%} & 3394 & 0.639 &  0.24 & 0.35 & {\bf 1.468} \\
	POS & -  & {\bf96.3\%} & 3394 & .64 &  .24 & .35 & {\bf 1.468} \\
	Z-forcing~\cite{goyal2017z}& $\checkmark$  & 47.7\% & \textbf{4361} & .79 & \textbf{.25} & .37 & 1.140 \\
	\specialcell{CVAE}& $\times$  & 12.1\% & 1991 & .52 & .16 & .29 & 0.959\\
	\specialcell{CVAE}&$\checkmark$  & 11.9\% & 1923 & .51 &.25 & .29 & 0.952\\
	\specialcell{Seq-CVAE+$\cN$}&$\times$  & 19.7\% & 2888 & .63 & .24 & .35 & 1.057\\
	\specialcell{Seq-CVAE+$\cN$}&$\checkmark$  & 52.8\% & 4162 & .69 & .25 & .43 & 1.244\\
	\specialcell{Seq-CVAE(BRNN)}&$\times$  & 91.8\% & 4267 & {.65} & {\bf .25} & {\bf .52}& 1.348\\
	\specialcell{Seq-CVAE}&$\checkmark$  & 94.0\% & 4266 & {\bf .52} & {\bf .25} & {\bf .54}& 1.448\\
% \midrule
% 	\multirow{5}{*}{\rotatebox[origin=c]{90}{\specialcell{Beam size \\\#samples: 100}}}
% 	&POS~\cite{deshpande2019fast} & 91.5\% &  3446 &  0.673 & 0.23 & 0.33 & 1.710\\
% 	&\specialcell{CVAE}  & 3.3\% & 2471 & - & .24 & .33 & 1.015\\
% 	&\specialcell{CVAE (ELMo)}  & 3.3\% & 2400 & - &.25 & .33 & 1.000\\
% 	&\specialcell{Seq-CVAE+$\cN$ Prior(ELMo)}  & 28.6\% & 4252 & - & .2 & .41 & 1.381\\
% 	&\specialcell{Seq-CVAE(BRNN)}  & 78.7\% & 4193 & 0.521 & 0.29 & 0.46 & 1.584\\
% 	&\specialcell{Seq-CVAE (ELMo)} & 84.2\% & \bf{4215} & \bf{0.640} & {\bf 0.33} & {\bf 0.48} & 1.695\\
	\specialrule{.1em}{.1em}{.1em}
	Human & -  & 99.8\% & - &  .510 &  .34 &  .48 & - \\
\bottomrule
\end{tabular}%
\caption{Diversity  and best-1 oracle accuracy on M-RNN test split for different models calculated using top-5 captions and consensus reranking. }
\label{tab:div-acc-abl}
\end{center}
\vspace{-0.8cm}
\end{table}

%% file: tables/ablation.tex
\begin{table}[t]
%\vspace{-2mm}
\centering
\footnotesize
\setlength\tabcolsep{1pt}
\begin{tabular}{lccccc}
\toprule
	Method & \elmo & Intention & Latent  & C@20 & C@100 \\ \\
	\toprule
	Seq-CVAE & $\times$ & $\cN$ & 512  & 1.015 $\pm$ 0.002 & 1.082 $\pm$ 0.001\\

\midrule
%	Seq-CVAE & \multirow{5}{*}{512}  & 1.016 $\pm$ 0.002 & 1.089 $\pm$ 0.002\\
%	\+ ELMO & & 1.332 $\pm$ 0.002 & 1.568 $\pm$ 0.004\\
%    \+ ELMO \& Shared LSTM & & 1.332 $\pm$0.002 & 1.573 $\pm$ 0.002 \\
	Seq-CVAE & $\times$ & $z_t | z_{<t}$ & \multirow{3}{*}{512}  & 1.016 $\pm$ 0.002 & 1.089 $\pm$ 0.002\\
	Seq-CVAE & $\checkmark$ & $z_t | z_{<t}$ & &1.332 $\pm$ 0.002 & 1.568 $\pm$ 0.004\\
    Seq-CVAE & $\checkmark$ & $z_t | z_{<t}, x_{<t}$ & &1.332 $\pm$ 0.002 & 1.573 $\pm$ 0.002 \\
    %Seq-CVAE & $\checkmark$ & $z_t | z_{<t}, x_{<t}$ & &$\checkmark$& \textbf{1.437} $\pm$ 0.002 & \textbf{1.686} $\pm$ 0.002 \\
\midrule
	Seq-CVAE & $\times$ & $z_t | z_{<t}$ & \multirow{3}{*}{256}  & 0.998 $\pm$ 0.014 & 1.059 $\pm$ 0.017\\
	Seq-CVAE & $\checkmark$ & $z_t | z_{<t}$ && 1.339 $\pm$ 0.003 &  1.559 $\pm$ 0.003\\
    Seq-CVAE & $\checkmark$ & $z_t | z_{<t}, x_{<t}$ && 1.335 $\pm$ 0.002 & 1.575 $\pm$ 0.004 \\
\midrule
	Seq-CVAE & $\times$ & $z_t | z_{<t}$ & \multirow{3}{*}{128}  & 0.957 $\pm$ 0.001 & 1.000 $\pm$ 0.002\\
    Seq-CVAE & $\checkmark$ & $z_t | z_{<t}$ & &1.328 $\pm$ 0.008  & 1.544 $\pm$ 0.006\\
    Seq-CVAE & $\checkmark$ & $z_t | z_{<t}, x_{<t}$ &  &1.324 $\pm$ 0.006 & 1.571 $\pm$ 0.002 \\
\bottomrule
\end{tabular}%

\caption{{\bf Ablation Analysis. }
%	We show an ablation study with different models, for latent dimensions of 128, 256 and 512. For each of these we observe that using the \elmo based representation improves the oracle CIDEr @100 by a considerable margin of $\sim$0.5 for all the 3 latent dimensions. Using \elmo along with a data dependent intention model gives the best performance, consistently achieving a CIDEr $> 1.57$. The low value of standard deviation calculated over 10 runs  for all the models is indicative that the learned latent space is robustly structured. Using a constant Gaussian  intention model ($\cN$) performs on par with using a parametric LSTM based intention model without \elmo, clearly showing the efficacy of the proposed approach.
	 We observe that  the \elmo based representation improves the oracle CIDEr @100 by $\sim$0.5. Using \elmo along with a data dependent intention model gives the best performance with CIDEr $\sim$1.573. The low value of the standard deviation calculated over 10 runs  for all the models is indicative that the learned latent space is robustly structured. Using a constant Gaussian  intention model ($\cN$) performs on par with using a parametric LSTM based intention model without \elmo, clearly showing the efficacy of the proposed approach.}
\label{tab:ablation}
\vspace{-.9cm}
\end{table}

%% file: fig/qualitative_figure.tex
\begin{figure}[t]
%\begin{figure}[H]
\centering
\scriptsize
\setlength{\tabcolsep}{2.0pt}
\resizebox{1\linewidth}{!}{
\begin{tabularx}{\columnwidth}{rXX}
		 \bf{$\alpha$} &\includegraphics[width=\linewidth,height=2.6cm]{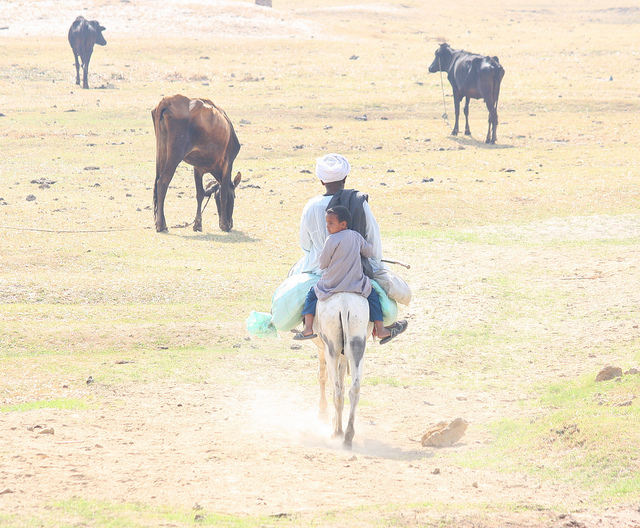}&
			\includegraphics[width=\linewidth,height=2.6cm]{fig/10707.jpg}\\
			 \cmidrule{1-3}
			\bf{0.0} &a horse is running down a dirt path  & a man and woman are playing video games together \\
			 %0.1 &a person riding a horse in a field & a white boat is traveling through the water & a man and woman are playing video games together\\
			 \bf{0.2} &a horse is being led by a man on a horse   & people are sitting on a couch with their feet up to the wii\\
			 %0.3 &a woman riding a horse while a dog is sitting on the ground & a boat is sailing in the water with a boat & a man and woman sitting on a couch with a wii remote\\
			 \bf{0.4} &a man riding a horse through a field &  a group of people are sitting on a couch with a wii remote in\\
			 %0.5 & a man riding a horse down the dirt road while people watch a dog & a boat is sailing in the water on a sunny day & people sitting on a couch with their laptops\\
			 \bf{0.6} &a woman walking across a dirt field with a horse &  people are sitting on a couch while playing a video game
			 \\
			 %0.7 &a woman is riding a horse in a field & a boat is sailing across the water with a sail boat in the background & people are sitting on a couch with their laptops \\
			 \bf{0.8} &a man riding a horse on a dirt field & a group of people are playing a video game \\
			 %0.9 &a woman is riding a horse on a dirt field & a boat is sailing on the water near a body of water& a group of people are sitting on a couch playing a video game\\
			\bf{1.0} &a woman riding a horse on a dirt field &  a group of people playing a video game together\\
\end{tabularx}}
\caption{Diversity of sentences controlled by linear interpolation between two samples. We observe meaningful sentences across all interpolated positions. Here $\alpha$ is the coefficient of linear interpolation.}
\label{fig:transition}
%\end{figure}
\vspace{-0.5cm}
\end{figure}

%% file: sections/conclude.tex
\vspace{\sectionReduceTop}
\section{Conclusion}
\label{sec:conclusion}
\vspace{\sectionReduceTop}
We propose \seqcvae{} which learns a word-wise latent space that captures the future of the sentence, \ie, the `intention' about how to complete the image description. This differs from existing techniques which generally learn a single latent space to  initialize sentence generation or to identically bias word generation  throughout the process. We demonstrate the proposed approach on the standard dataset and illustrate results on par \wrt baseline accuracies while significantly improving a large variety of diversity metrics.
\\
{\bf Acknowledgments.} Supported in part by  NSF grant 1718221, Samsung, and 3M. We thank NVIDIA for GPUs. % used in this work.
%\vspace{\sectionReduceBot}